\documentclass[manuscript,screen]{acm/acmart}

\usepackage[T1]{fontenc}
\usepackage[utf8]{inputenc}
\usepackage[edges]{forest}
\usepackage{microtype}

\usepackage{graphicx}
\usepackage[most,skins,theorems]{tcolorbox}
\usepackage{color}
\usepackage{times}
\usepackage{latexsym}
\usepackage{adjustbox}
\usepackage{xspace}
\usepackage{todonotes}
\usepackage{xcolor}
\usepackage{comment}
\usepackage{tikz}
\usepackage{pgfplots}
\usepackage{pgfplotstable}
\usepackage{makecell}
\usepackage{adjustbox}
\usepackage{pifont}
\usepackage{float}
\usepackage{placeins}
\usepackage[edges]{forest}
\definecolor{hidden-draw}{RGB}{0,0,0}
\usepackage{multirow}
\usepackage{hyperref}
\usepackage[all]{hypcap}

\usepackage{tcolorbox}
\usepackage{amsmath}
\makeatletter
\@ifundefined{Bbbk}{\usepackage{amssymb}}{}
\makeatother

\usepackage{booktabs}
\usepackage{threeparttable}
\usepackage{tabularx}

\usepackage{soul}
\usepackage{xcolor}
\usepackage{subfig}

\usepackage{bm} %

\usepackage{algorithm}
\usepackage{algorithmic}

\usepackage{wrapfig}

\definecolor{mygreen}{RGB}{11,141,10}
\definecolor{myred}{RGB}{223,68,52}
\definecolor{myblue}{RGB}{70,130,180}
\definecolor{mydeepblue}{RGB}{65,105,225}
\definecolor{myviolet}{RGB}{97,0,138}
\definecolor{myburgundy}{RGB}{110,10,30}
\definecolor{myblue2}{RGB}{0,105,148}
\definecolor{iceblue}{RGB}{173, 216, 230}
\definecolor{puregreen}{RGB}{0, 70, 0}

\definecolor{wingreen}{rgb}{0,0.45,0.24}
\definecolor{losered}{rgb}{1.0,0.1,0.24}

\definecolor{lightcoral}{rgb}{0.97, 0.36, 0.46}
\definecolor{lightyellow}{rgb}{0.98, 0.7, 0}
\definecolor{harvestgold}{rgb}{0.85, 0.57, 0.0}
\definecolor{brightlavender}{rgb}{0.75, 0.58, 0.89}
\definecolor{capri}{rgb}{0.0, 0.75, 1.0}
\definecolor{carminepink}{rgb}{0.92, 0.3, 0.26}
\definecolor{celadon}{rgb}{0.67, 0.88, 0.69}
\definecolor{darkpastelgreen}{rgb}{0.01, 0.75, 0.24}

\definecolor{grayhighlight}{RGB}{250,250,227}

\definecolor{target}{HTML}{F47983}
\definecolor{control}{HTML}{3E87CD}
\definecolor{credibility}{HTML}{B98AC9}
\definecolor{logical}{HTML}{93C572}
\definecolor{emotional}{HTML}{F9EAC3}

\usepackage{pifont}

\usepackage[T1]{fontenc}
\usepackage{hyperref}
\usepackage[utf8]{inputenc}
\usepackage{wrapfig}
\usepackage{microtype}
\usepackage{tikz} 
\usepackage{makecell}
\usetikzlibrary{shapes, arrows.meta, positioning, shadows, backgrounds, fit}
\usepackage{booktabs} % 如果你配合三线表使用
\usepackage{arydshln} % 必须放在 booktabs 之后加载
\usepackage{booktabs}   % 画三线表
\usepackage{tabularx}   % 自动调整列宽
\usepackage{multirow}   % 合并单元格
\usepackage{pifont}     % 用于对勾符号 (\ding{51})
\usepackage{xcolor}     % 用于颜色 (可选)

\newcommand{\cmark}{\textcolor{green!60!black}{\ding{51}}} % 绿色对勾
\newcommand{\xmark}{\textcolor{red!60!black}{\ding{55}}}   % 红色叉号
\tcbset{
  aibox/.style={
    width=\linewidth,
    top=8pt,
    bottom=4pt,
    colback=blue!6!white,
    colframe=black,
    colbacktitle=black,
    enhanced,
    center,
    attach boxed title to top left={yshift=-0.1in,xshift=0.15in},
    boxed title style={boxrule=0pt,colframe=white,},
  }
}
\newtcolorbox{AIbox}[2][]{aibox,title=#2,#1}

\NewDocumentCommand{\hr}
{ mO{} }{\textcolor{purple}{\textsuperscript{\textit{Hongru}}\textsf{\textbf{\small[#1]}}}}

\NewDocumentCommand{\by}
{ mO{} }{\textcolor{blue}{\textsuperscript{\textsl{Boyang}}{\textbf{\small[#1]}}}}

\NewDocumentCommand{\jh}
{ mO{} }{\textcolor{cyan}{\textsuperscript{\textsl{Jianhui}}{\textbf{\small[#1]}}}}

\usetikzlibrary{shapes.geometric, arrows.meta, positioning, calc, decorations.pathreplacing, shadows}

\definecolor{goodGreen}{RGB}{82, 189, 149}
\definecolor{badRed}{RGB}{231, 111, 81}
\definecolor{neutralBlue}{RGB}{42, 157, 143}
\definecolor{lightGray}{RGB}{240, 240, 240}

\AtBeginDocument{%
  }

\graphicspath{{acm/}}

\setcopyright{acmlicensed}
\copyrightyear{2018}
\acmYear{2018}
\acmDOI{XXXXXXX.XXXXXXX}
\acmConference[Conference acronym 'XX]{Make sure to enter the correct
  conference title from your rights confirmation email}{June 03--05,
  2018}{Woodstock, NY}
\acmISBN{978-1-4503-XXXX-X/2018/06}

\begin{document}

\title{Harnessing the Reasoning Economy: A Survey of Efficient Reasoning for Large Language Models}

\author{Rui Wang}
\authornote{Project Lead and Equal Contributions.}
\affiliation{%
  \institution{The Chinese University of Hong Kong}
  \country{Hong Kong SAR, China}
}
\email{rwang@se.cuhk.edu.hk}

\author{Hongru Wang}
\authornotemark[1]
\affiliation{%
  \institution{University of Edinburgh}
  \country{Edinburgh, UK}
}
\email{hrwang@se.cuhk.edu.hk}

\author{Boyang Xue}
\authornotemark[1]
\affiliation{%
  \institution{The Chinese University of Hong Kong}
  \country{Hong Kong SAR, China}
}
\email{byxue@se.cuhk.edu.hk}

\author{Yixia Li}
\authornote{Significant Contributions.}
\affiliation{%
  \institution{Southern University of Science and Technology}
  \country{China}
}

\author{Jianhui Pang}
\authornotemark[2]
\affiliation{%
  \institution{University of Macau}
  \country{Macau SAR, China}
}
\email{nlp2ct.pangjh3@gmail.com}

\author{Shudong Liu}
\authornotemark[2]
\affiliation{%
  \institution{University of Macau}
  \country{Macau SAR, China}
}
\email{nlp2ct.shudong@gmail.com}

\author{Yi Chen}
\affiliation{%
  \institution{The University of Hong Kong}
  \country{Hong Kong SAR, China}
}

\author{Jiahao Qiu}
\affiliation{%
  \institution{Princeton University}
  \country{USA}
}

\author{Derek Fai Wong}
\affiliation{%
  \institution{University of Macau}
  \country{Macau SAR, China}
}

\author{Guanhua Chen}
\affiliation{%
  \institution{Southern University of Science and Technology}
  \country{China}
}

\author{Heng Ji}
\affiliation{%
  \institution{University of Illinois Urbana-Champaign}
  \country{USA}
}

\author{Kam-Fai Wong}
\affiliation{%
  \institution{The Chinese University of Hong Kong}
  \country{Hong Kong SAR, China}
}

\renewcommand{\shortauthors}{Wang et al.}

\begin{abstract}
  Recent advancements in Large Language Models (LLMs) have significantly enhanced their ability to perform complex reasoning tasks, transitioning from fast and intuitive thinking (System 1) to slow and deep reasoning (System 2).
  While System 2 reasoning improves task accuracy, it often incurs substantial computational costs due to its slow thinking nature and inefficient or unnecessary reasoning behaviors.
  In contrast, System 1 reasoning is computationally efficient but leads to suboptimal performance.
  Consequently, it is critical to balance the trade-off between performance (\textit{benefits}) and computational costs (\textit{budgets}), giving rise to the concept of \textit{reasoning economy}.
In this survey, we provide a comprehensive analysis of the reasoning economy by structuring the landscape into a lifecycle-based taxonomy:
(1) Sources of Inefficiency: We first analyze the mechanisms of reasoning inefficiency, defining where the static allocation of computation fails to align with dynamic task complexity.
(2) Post-training Optimization (Reasoning Capabilities Supply-side): We review methods for constructing high-density reasoning assets, focusing on building models with inherent efficiency before deployment.
(3) Inference Optimization (Reasoning Capabilities Demand-side): We explore dynamic system designs for adaptive budget management, where computational resources are allocated on-demand to maximize the return on inference costs.
We conclude that advancing current LRMs requires a synergistic approach: effective post-training methods to enhance \textbf{intrinsic per-token intelligence}, coupled with adaptive test-time strategies to fully \textbf{unlock model potential}. This dual optimization is essential to approach performance ceilings while minimizing computational costs.
By offering insights and highlighting open challenges, we aim to shed light on strategies for improving the reasoning economy of LLMs, thereby serving as a valuable resource for advancing research in this evolving area\footnote{\href{https://github.com/DevoAllen/Awesome-Reasoning-Economy-Papers}{https://github.com/DevoAllen/Awesome-Reasoning-Economy-Papers}}.
  % In this survey, we provide a comprehensive analysis of reasoning economy in both the post-training and test-time inference stages of LLMs, encompassing i) the cause of reasoning inefficiency, ii) behavior analysis of different reasoning patterns, and iii) potential solutions to achieve reasoning economy.
  % By offering actionable insights and highlighting open challenges, we aim to shed light on strategies for improving the reasoning economy of LLMs, thereby serving as a valuable resource for advancing research in this evolving area.
  % We also provide a repository to continually track developments in this fast-evolving field\footnote{\href{https://github.com/DevoAllen/Awesome-Reasoning-Economy-Papers}{https://github.com/DevoAllen/Awesome-Reasoning-Economy-Papers}}.
\end{abstract}



\maketitle

% \begin{figure*}
%     \centering
%     \includegraphics[width=0.8\linewidth]{Figs/teaser.pdf}
%     \caption{The Landscape of Reasoning Economy}
%     \label{fig:teaser}
% \end{figure*}
\begin{wrapfigure}{r}{0.5\textwidth} 
    \centering
    \includegraphics[width=\linewidth]{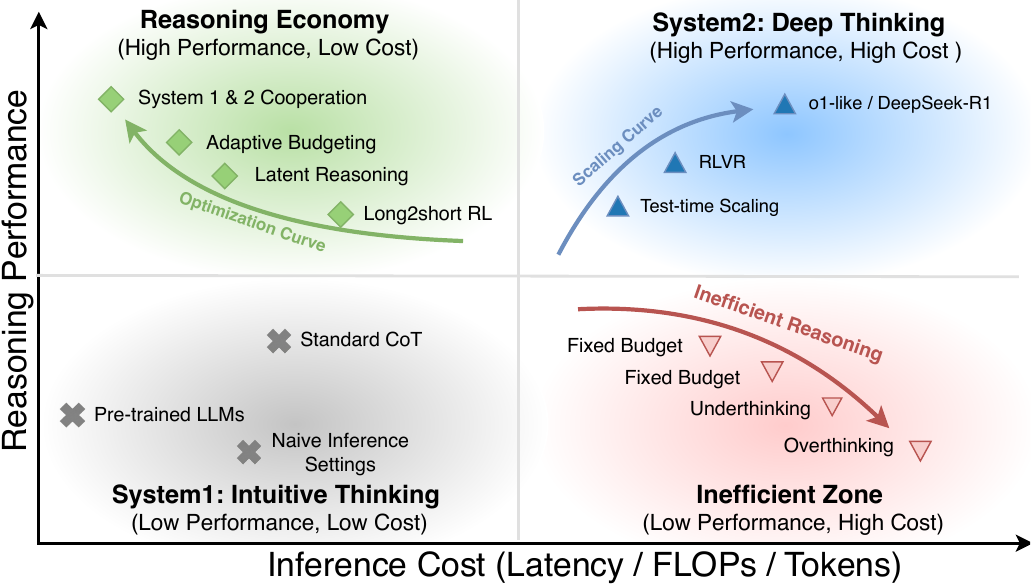}
    \caption{The Landscape of Reasoning Economy}
    \label{fig:teaser}
    \vspace{-10pt} 
\end{wrapfigure}

\section{Introduction}

Large Language Models (LLMs) have demonstrated exceptional performance across various language understanding and generation tasks, particularly with the advent of Chain-of-Thought (CoT) prompting \citep{cot}, which encourages models to generate explicit, step-by-step reasoning to arrive at the final answers.
While LLMs excel in many scenarios, their reliance on fast, intuitive thinking often falls short when faced with complex reasoning challenges, such as advanced mathematics \cite{aime, btm-1} and coding tasks.
Therefore, recent studies attempt to further boost the reasoning capabilities of LLMs, as exemplified by OpenAI o1 \cite{o1}, DeepSeek R1 \cite{deepseek-r1} and QwQ \cite{noauthor_qwq-32b_nodate}, to thoroughly explore potential solutions and improve problem-solving accuracy with slow and deep thinking~\cite{hongruself}, leading to the blossoming of Large Reasoning Models (LRMs) and new scaling law during the inference \cite{snell_scaling_2024}.

However, this cognitive upgrade comes with a prohibitive price tag. 
The deliberative nature of LRMs incurs substantial inference overhead due to the generation of elongated CoT sequences.
A critical bottleneck in current deployments is the static, \textit{one-size-fits-all} strategy: heavy reasoning is applied indiscriminately, regardless of task difficulty. 
This results in severe \textbf{resource misalignment}, where models waste computation on trivial queries (\textit{overthinking})~\cite{over-cau-1,over-cau-2-agent} while failing to allocate sufficient depth to truly intricate problems (\textit{underthinking})~\cite{snell_scaling_2024,decp-behavir-1-underthink}.
% Empirical evidence suggests that not all generated tokens contribute significantly to the final answer, highlighting a massive structural inefficiency in current systems.
This imbalance between capability and efficiency presents a significant challenge - achieving \textit{\textbf{Reasoning Economy}}, a global optimum that optimizes token usage (\textit{\textbf{budgets}}) by emphasizing meaningful reasoning steps, reducing redundancy, and dynamically adjusting computational effort based on task complexity.
Therefore, it not only reduces computation and latency of LRMs but also unlocks their potential by intelligently stopping or diving deeper (\textit{\textbf{benefits}})~\cite{early-stop-2,team_kimi_2025}.
With the growing importance of the reasoning economy, there is an urgent need to systematically understand and analyze different reasoning behaviors of LRMs, reveal potential challenges toward efficient LRMs, and clearly showcase solutions to achieve the reasoning economy.

\begin{figure*}
    \centering
    \includegraphics[width=1\linewidth]{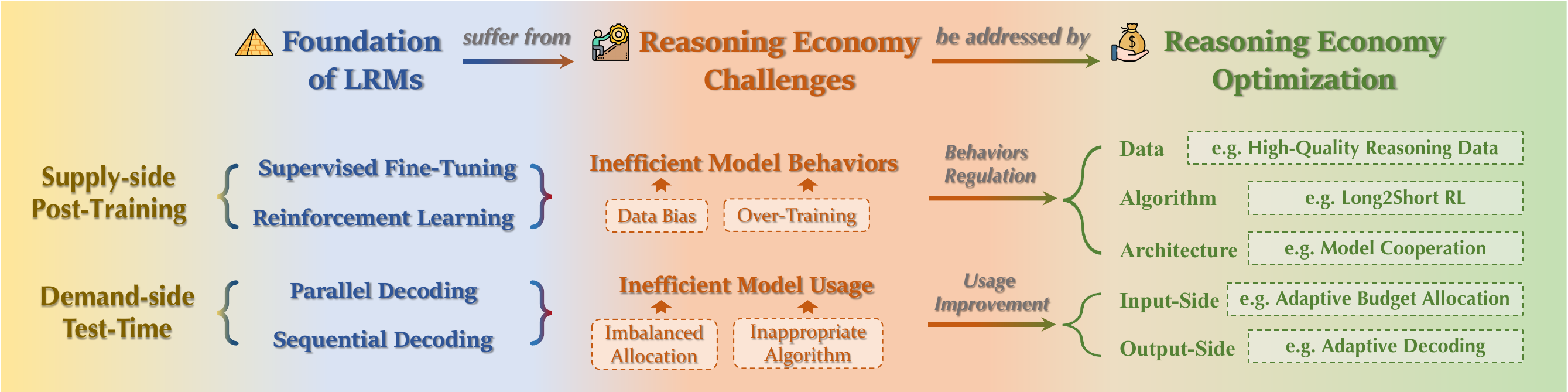}
    \caption{We first provide a review of the Foundation of LRMs, including post-training and test-time methods. Then, we analyze the Challenges of the Reasoning Economy and survey the efforts to achieve the reasoning economy by regulating inefficient model behaviors and improving test-time budget allocation and utilization.}
    \label{fig:method}
\end{figure*}

\subsection{Review Structure}
In this survey, we present the first comprehensive and systematic review of reasoning economy for LRMs.
Specifically, we begin by establishing the \textit{Foundation of LRMs} (\S~\ref{sec:basic-r-llms}), quickly dissecting how post-training methods (\S~\ref{subsec: post-train}) (i.e, supervised fine-tuning and reinforcement learning) shape reasoning behaviors and how test-time strategies (\S~\ref{subsec: test-time-methods}) (i.e, parallel and sequential) influence the model performance.
Building on this foundation, we conduct a rigorous analysis of \textit{Challenges towards Reasoning Economy} (\S~\ref{sec:challeng}), classifying them into inefficient behaviors from the model itself (\S~\ref{subsec:ineff-pt}) or inefficient usage during test-time (\S~\ref{subsec:ineff-tt}).
% distinct categories based on their underlying causes and patterns.
Finally, we discuss potential solutions for optimizing reasoning economy in terms of two directions: i) constructing high-efficiency assets by behaviors regulations in post-training(\S~\ref{sec:solu-pt}), which targets undesirable reasoning behaviors at their source in terms of data, algorithm and even model architecture, 
and ii) dynamic budget management systems in test-time (\S~\ref{sec:solu-tt}) to dynamically adapt computation. We also discuss several open challenges, and suggest future research directions (\S~\ref{sec:dis}). By offering a clear and structured roadmap, our work aims to provide actionable insights to guide future research and foster the development of the reasoning economy for more sustainable LRMs.

\subsection{Position and Contribution}

Recent surveys have extensively covered the construction~\cite{S1-S2-survey} and mechanisms~\cite{chen2025reasoningerasurveylong} of LRMs. 
Concurrently, recognizing the computational costs associated with deep reasoning, several works~\cite{ot-infer-survey-1, ot-survey-no-issuse,ot-survey-1,ot-survey-2} have shifted focus toward \textit{efficient reasoning}.
% Given the escalating costs of deep reasoning, distinct attention has shifted toward \textit{efficient reasoning}~\cite{ot-infer-survey-1, ot-survey-no-issuse}. 
Unlike system-level surveys focusing on pruning or distributed computing~\cite{zhou2024surveyefficientinferencelarge,llm-prune-1}, this line of work aligns with our focus on \textbf{reasoning efficiency}, optimizing the allocation of computational budgets during the reasoning process.
However, despite this progress, current literature remains limited in three critical dimensions:

\noindent \textbf{(1) Superficial Phenomenon Analysis:} Prior works~\cite{ot-infer-survey-1, ot-survey-no-issuse} primarily concentrate on catalog symptoms of inefficiency, and methods to alleviate them, without dissecting their underlying \textit{causes}, e.g.., data bias~\cite{singhal2024a,RL-scaling} or reward hacking~\cite{reward-hack-1}.

\noindent \textbf{(2) Narrow Optimization Goals:} Most surveys~\cite{ot-survey-1,ot-survey-2} focus narrowly on \textit{compression}, neglecting the broader paradigm of \textit{dynamic allocation}, where efficiency implies investing \textit{more} budget on hard tasks and \textit{less} on easy ones.

\noindent \textbf{(3) Fragmented Perspectives:} Existing research often isolates training from inference~\cite{ot-infer-survey-1}. There lacks a unified framework that views reasoning efficiency as a \textbf{lifecycle problem}, bridging the production of intelligence (\textit{Supply}) with its adaptive consumption (\textit{Demand}).

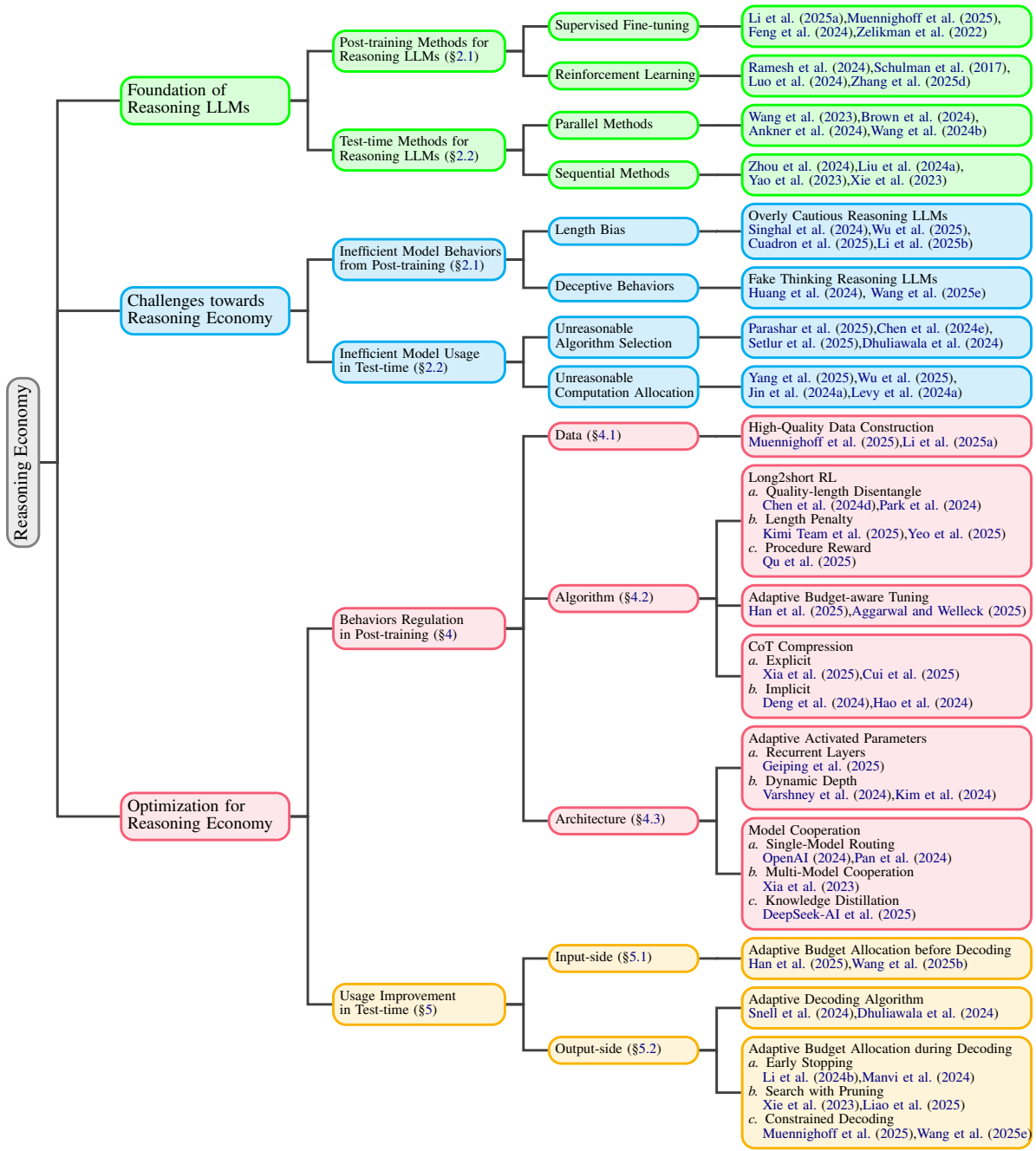
\begin{figure*}[t!]
    \centering
    \tikzstyle{my-box}=[
    rectangle,
    rounded corners,
    text opacity=1,
    minimum height=0.1em,
    minimum width=0.1em,
    inner sep=2pt,
    align=left,
    fill opacity=.5,
    ]
    \tikzstyle{leaf}=[my-box]
    \resizebox{0.97\textwidth}{!}{
        \begin{forest}
            forked edges,
            for tree={
                grow=east,
                reversed=true,
                anchor=base west,
                parent anchor=east,
                child anchor=west,
                base=left,
                font=\fontsize{6}{6}\selectfont,
                rectangle,
                draw=hidden-draw,
                rounded corners,
                align=left,
                minimum width=0.1em,
                edge+={darkgray, line width=0.8pt},
                s sep=2pt,
                inner xsep=2pt,
                inner ysep=2pt,
                ver/.style={rotate=90, child anchor=north, parent anchor=south, anchor=center},
            },
            where level=1{text width=5.0em,font=\fontsize{5.5}{5.5}\selectfont,}{},
            where level=2{text width=5.6em,font=\fontsize{4.5}{4.5}\selectfont}{},
            where level=3{text width=4.9em,font=\fontsize{4.5}{4.5}\selectfont,}{},
            where level=4{text width=9.2em,font=\fontsize{4.5}{4.5}\selectfont,}{},
            where level=5{text width=9.2em,font=\fontsize{4.5}{4.5}\selectfont,}{},
            [
                Reasoning Economy, draw=gray, color=gray!100, fill=gray!15, thick, text=black,ver
                [
                    Foundation of \\ Reasoning LLMs
                    , color=green!100, fill=green!15, thick, text=black
                    [
                        {Post-training} Methods for \\ Reasoning LLMs (\S\ref{subsec: post-train})
                        , color=green!100, fill=green!15, thick, text=black
                        [
                            Supervised Fine-tuning
                        , color=green!100, fill=green!15, thick, text=black
                            [
                                \citet{limr}{,}\citet{force-decode-s1}{,}\\
                                \citet{feng-etal-2024-improving}{,}\citet{zelikman2022star}
                                , color=green!100, fill=green!15, thick, text=black
                            ]
                        ]
                        [
                           Reinforcement Learning
                        , color=green!100, fill=green!15, thick, text=black
                            [
                                \citet{grpo}{,}\citet{ppo}{,}\\
                                \citet{luo2024improve}{,}\citet{zhang2025lessons}
                                , color=green!100, fill=green!15, thick, text=black
                            ]
                        ]
                    ]
                    [
                        Test-time Methods for \\ Reasoning LLMs (\S\ref{subsec: test-time-methods})
                        , color=green!100, fill=green!15, thick, text=black
                        [
                            Parallel Decoding
                            , color=green!100, fill=green!15, thick, text=black
                            [
                                \citet{selfconsistency}{,}\citet{brown2024largelanguagemonkeysscaling}{,}\\
                                \citet{ovm-1}{,}\citet{wang_math-shepherd_2024}
                                , color=green!100, fill=green!15, thick, text=black
                            ]
                        ]
                        [
                           Sequential Decoding
                            , color=green!100, fill=green!15, thick, text=black
                            [
                                \citet{self-refine}{,}\citet{shi2025ssrsocraticselfrefinelarge}{,}\\
                                \citet{xu2025doublecheckerenhancingreasoningslowthinking}{,}\citet{tie2025llmscorrectthemselvesbenchmark}
                                , color=green!100, fill=green!15, thick, text=black
                            ]
                        ]
                        [
                           Hybrid Decoding
                            , color=green!100, fill=green!15, thick, text=black
                            [
                                \citet{guided-beam-search-2}{,}\citet{ovm-guided-beam-search-1}{,}\\
                                \citet{tot}{,}\citet{mcts-1}
                                , color=green!100, fill=green!15, thick, text=black
                            ]
                        ]
                    ]
                ]
                [
                    Challenges towards \\ Reasoning Economy
                    , color=cyan!100, fill=cyan!15, thick, text=black
                    [
                        Inefficient Model Behaviors \\
                        from Post-training (\S\ref{subsec: post-train})
                        , color=cyan!100, fill=cyan!15, thick, text=black
                        [
                            Overly Cautious
                            , color=cyan!100, fill=cyan!15, thick, text=black
                            [
                                {\citet{singhal2024a}}{,}{\citet{over-cau-1}}{,}\\
                                {\citet{over-cau-2-agent}}{,}\citet{S1-S2-survey}
                                , color=cyan!100, fill=cyan!15, thick, text=black
                            ]
                        ]
                        [
                            Superficial Reasoning
                            , color=cyan!100, fill=cyan!15, thick, text=black
                            [
                                \citet{self-critique-bc-2}{,}
                                \citet{decp-behavir-1-underthink}
                                , color=cyan!100, fill=cyan!15, thick, text=black
                            ]
                        ]
                    ]
                    [
                        Inefficient Budget Allocation \\
                        in Test-time (\S\ref{subsec: test-time-methods})
                        , color=cyan!100, fill=cyan!15, thick, text=black
                        [
                            Unreasonable \\ Algorithm Selection
                            , color=cyan!100, fill=cyan!15, thick, text=black
                            [
                                \citet{not-alog-is-all}{,}\citet{vote-lead-to-worse}{,}\\
                                \citet{veri-needed-setlur_scaling_2025}{,}\citet{daptive_temp}
                                ,  color=cyan!100, fill=cyan!15, thick, text=black
                            ]
                        ]
                        [
                            Unreasonable \\ Computation Allocation
                            , color=cyan!100, fill=cyan!15, thick, text=black
                            [
                                \citet{domain-optimal-len}{,}\citet{over-cau-1}{,}\\
                                \citet{longer-is-not-better}{,}\citet{length-analysis-on-llms-1}
                                ,  color=cyan!100, fill=cyan!15, thick, text=black
                            ]
                        ]
                    ]
                ]
                [
                    Optimization for \\ Reasoning Economy
                    , color=lightcoral!100, fill=lightcoral!15, thick, text=black
                    [
                        Behaviors Regulation \\ in Post-training (\S\ref{sec:solu-pt})
                        , color=lightcoral!100, fill=lightcoral!15, thick, text=black
                        [
                            Data (\S\ref{subsec:solu-pt-data})
                            , color=lightcoral!100, fill=lightcoral!15, thick, text=black
                            [
                                High-Quality Data Construction\\
                                \citet{force-decode-s1}{,}\citet{limr}
                                , color=lightcoral!100, fill=lightcoral!15, thick, text=black
                            ]
                        ]
                        [
                            Algorithm (\S\ref{subsec:solu-pt-alg})
                            , color=lightcoral!100, fill=lightcoral!15, thick, text=black
                            [
                                {Long2short RL}\\
                                {\textit{a. }Quality-length Disentangle }\\
                                \quad \citet{chen2024odin}{,}\citet{park-etal-2024-disentangling} \\
                                {\textit{b.} Length Penalty }\\
                                \quad \citet{team_kimi_2025}{,}\citet{yeo_demystifying_2025} \\
                                {\textit{c.} Procedure Reward }\\
                                \quad \citet{qu_MRL_optimizing_2025}
                                , color=lightcoral!100, fill=lightcoral!15, thick, text=black
                            ]
                            [
                                Adaptive Budget-aware Tuning \\
                               {\textit{a.} Budget Specification }\\
                               \quad \citet{wen2025budgetthinkerempoweringbudgetawarellm}{,}\citet{zhao2025saberswitchablebalancedtraining} \\
                               {\textit{b.} Budget-aligned Post-training }\\
                               \quad \citet{yuan2024-length-cons-sft}{,}\citet{lyu2025hierarchicalbudgetpolicyoptimization} \\
                               {\textit{c.} Adaptive Allocation and Gating }\\
                               \quad \citet{yan2025drqadynamicreasoningquota}{,}\citet{jiang2025think}
                                , color=lightcoral!100, fill=lightcoral!15, thick, text=black
                            ]
                            [
                                CoT Compression \\
                               {\textit{a.} Explicit }\\
                               \quad \citet{xia2025tokenskip}{,}\citet{cui2025stepwise} \\
                               {\textit{b.} Implicit }\\
                               \quad \citet{deng2024explicitcotimplicitcot}{,}\citet{hao2024traininglargelanguagemodels}
                                , color=lightcoral!100, fill=lightcoral!15, thick, text=black
                            ]
                        ]
                        [
                            Architecture (\S\ref{subsec:solu-pt-arc})
                            , color=lightcoral!100, fill=lightcoral!15, thick, text=black
                            [
                                Adaptive Activated Parameters \\
                               {\textit{a.} Recurrent Layers}\\
                               \quad \citet{geiping2025scalingtesttimecomputelatent}\\
                               {\textit{b.} Dynamic Depth}\\
                               \quad \citet{varshney-etal-2024-investigating}{,}\citet{kim2024shortenedllamadepthpruning}
                                , color=lightcoral!100, fill=lightcoral!15, thick, text=black
                            ]
                            [
                                Model Cooperation\\
                               {\textit{a.} Single-Model Routing}\\
                               \quad \citet{o1}{,}\citet{pan-etal-2024-dynathink}\\
                               {\textit{b.} Multi-Model Cooperation}\\
                               \quad \citet{xia-etal-2023-speculative}{,}\citet{wu2026faircomprehensiveevaluationrouters} \\
                               {\textit{c.} Knowledge Distillation}\\
                               \quad \citet{deepseek-r1}
                                , color=lightcoral!100, fill=lightcoral!15, thick, text=black
                            ]
                        ]
                    ]
                    [
                        Adaptive Budgeting \\ in Test-time (\S\ref{sec:solu-tt})
                        , color=lightyellow!100, fill=lightyellow!15, thick, text=black
                        [
                            Input-side (\S\ref{subsec:solu-tt-input})
                            , color=lightyellow!100, fill=lightyellow!15, thick, text=black
                            [
                               Adaptive Budget Allocation before Decoding\\
                               \citet{0shot-sft-length-1}{,}\citet{difficulty-aware-budget-pred}
                               , color=lightyellow!100, fill=lightyellow!15, thick, text=black
                            ]
                        ]
                        [
                            Output-side (\S\ref{sec:solu-tt-output})
                            , color=lightyellow!100, fill=lightyellow!15, thick, text=black
                            [
                                Adaptive Decoding Algorithm \\
                                \citet{snell_scaling_2024}{,}\citet{daptive_temp}
                                , color=lightyellow!100, fill=lightyellow!15, thick, text=black
                            ]
                            [
                               Adaptive Budget Allocation during Decoding \\
                               {\textit{a.} Early Stopping} \\
                               \quad \citet{early-stop-3}{,}\citet{early-stop-1} \\
                               {\textit{b.} Search with Pruning} \\
                               \quad \citet{guided-beam-search-2}{,}\citet{liao2025rewardguidedspeculativedecodingefficient}\\
                               {\textit{c.} Constrained Decoding} \\
                               \quad \citet{force-decode-s1}{,}\citet{decp-behavir-1-underthink}
                               , color=lightyellow!100, fill=lightyellow!15, thick, text=black
                            ]
                        ]
                    ]
                ]
            ]
        \end{forest}

        }
    \caption{Taxonomy of achieving Reasoning Economy, from foundation, challenges, to solutions.}
    \label{fig:taxonomy}
\end{figure*}
\FloatBarrier

% \clearpage

\section{Foundations of LRMs} \label{sec:basic-r-llms}

This section outlines the fundamental methodologies that underpin the advanced reasoning capabilities of LRMs.
We categorize these foundations into two stages:
\textbf{Post-training} (\S\ref{subsec: post-train}), which shapes the intrinsic capabilities and behaviors of the model (the \textit{supply} of intelligence);
and \textbf{Test-time Scaling} (\S\ref{subsec: test-time-methods}), which determines how these capabilities are utilized to approach performance ceilings through computational scaling (the \textit{demand} management).

\subsection{Post-training} \label{subsec: post-train}

Post-training is the critical phase where pre-trained base models are transformed into instruction-following reasoning agents. This process typically follows a two-stage pipeline: \textbf{Supervised Fine-Tuning (SFT)} to establish behavioral patterns \cite{flan-sft,instructgpt}.
Building upon this foundation, \textbf{Reinforcement Learning (RL)} is then utilized to optimize the reasoning logic itself \cite{rlhf,ppo,grpo}.
Crucially, RL enables the model to look beyond the training data, utilizing reward signals to explore diverse reasoning paths and self-correct errors, which is essential for aligning with rigorous logic and complex human preferences \cite{deepseek-r1,o1}.
% , and \textbf{Reinforcement Learning (RL)} to optimize reasoning logic and align with human preferences \cite{rlhf,ppo,grpo}.

% \lgnote{Post-training Methods for LRMs}

\paragraph{Supervised Fine-tuning}
SFT plays a crucial role in enhancing the zero-shot multi-task performance of LLMs \cite{achiam2023gpt} and serves as the foundational phase for initializing reasoning capabilities, a process often characterized as behavioral cloning. 
Consequently, the primary objective of the SFT stage is to internalize desired reasoning patterns by curating and incorporating target model behaviors into high-quality demonstration datasets.
By leveraging high-quality task-specific data \cite{alpaca,lima}, SFT refines the model’s ability to generalize across various domains, improving both accuracy and reliability in real-world applications, such as summarization \cite{fabbri2021summeval}, machine translation \cite{pang2025salute}, and question answering \cite{kwiatkowski-etal-2019-natural}.
Recent studies have proposed leveraging self-improvement methodologies to enhance the reasoning capabilities of models
\cite{zelikman2022star,huang-etal-2023-large,gulcehre2023reinforced,khattab2023dspy,zelikman2024quiet,feng-etal-2024-improving,li2024large,subramaniam2025multiagent}.
STaR \cite{zelikman2022star} employs an iterative approach where the large LLM is prompted to generate multiple reasoning chains until the correct solution is obtained, after which the model is fine-tuned on the complete set of successful reasoning trajectories. The latest work Self-Reasoning Language Models (SRLM)~\citep{hongruself} further extends to more general instruction-tuning datasets by mixing with few reasoning catalyst data, enabling SRLM to self-refine its own rationales iteratively.
Though \citet{deepseek-r1} proves that the SFT stage is no longer a must for LRMs by performing RL for pre-trained model,
they also found that SFT contributes to accelerating the training process and achieving better performance for LRMs.

\paragraph{Reinforcement Learning}

In the training of LRMs, Reinforcement Learning (RL) plays a key role in enhancing LLMs' reasoning capabilities by providing rewards for both the underlying reasoning process and the final answer \cite{o1,deepseek-r1,ppo,grpo}, rather than relying exclusively on token-by-token supervision. Moreover, recent studies \cite{sft_memo_RL_gene} further indicate that RL-based approaches significantly enhance the generalizability of LLMs, allowing them to perform better in a wider range of tasks and scenarios. These findings underscore the importance of RL in refining model adaptability and improving reasoning capabilities.
The core focus of reinforcement learning currently lies in the design of the reward signals.
According to the different granularity of reward signals, there are two major reward models: the Process Reward Model (PRM) and the Outcome Reward Model (ORM).
Specifically, the PRM \cite{prm-1} assigns rewards based on intermediate steps within an action sequence rather than solely on the final outcome \cite{zhang2025lessons}.
PRM enables more fine-grained learning signals, guiding the LLMs toward optimal policies by rewarding beneficial intermediate behaviors.
However, there are several limitations.
One the one hand, the training data of PRM is challenging to obtain, as it either requires extensive human annotations \cite{prm-1} or a large amount of sampling \cite{wang_math-shepherd_2024}, which limits its application.
On the other hand, they may also be overly stringent for the reasoning capabilities of LLMs \cite{sft_memo_RL_gene}.
% Overall, the rewards assigned by the PRM can act as a strong guiding signal for the reasoning process of LRMs.

% \hr{please add some citations here for reward hacking and others.}

% In contrast, the ORM \cite{gsm8k} assigns rewards based on the final outcome of solutions \cite{ovm-1,ovm-2,ovm-3,luo2024improve}.
% ORM is widely used in tasks like mathematical reasoning and decision-making, as Reinforcement Learning with Verifiable Rewards (RLVR), where the final result could be explicitly given and evaluated.
% Therefore, it is relatively easier to directly assign the reward based on some rule-based method such as format and answer matching, as used in \citet{deepseek-r1} and \citet{reft}.
% Despite ORM only provides a supervision signal at the outcome level, it still brings exceptional reasoning capabilities since it allows language models to explore reasoning paths without restrictive constraints.
% For example, the R1 model exhibits the "Aha" moment \cite{deepseek-r1}, i.e., the emerged self-refine and critique abilities with only rule-based accuracy reward.
% In summary, both PRM and ORM offer distinct advantages and limitations in training LRMs.
% More effective reward modeling combining the strengths of PRM and ORM still remains an open question.
In contrast, Outcome Reward Models (ORMs) \cite{gsm8k} evaluate reasoning chains solely based on the correctness of the final result \cite{ovm-1,ovm-2,ovm-3,luo2024improve}. This approach is foundational to \textit{Reinforcement Learning with Verifiable Rewards} (RLVR), particularly in deterministic domains like mathematics and coding, where ground truth is easily verifiable. 
Consequently, ORMs facilitate scalable training via rule-based outcome reward functions (e.g., format and answer matching), as utilized in \citet{deepseek-r1} and \citet{reft}. 
Although ORMs provide sparse supervision signals, they foster exceptional reasoning capabilities by allowing the model to freely explore the solution space without imposing rigid constraints on the reasoning process. 
A notable example is the DeepSeek-R1 model, which demonstrated emergent "Aha moments", spontaneous self-refinement and error correction, driven purely by rule-based accuracy rewards \cite{deepseek-r1}. 

In summary, the choice between PRMs and ORMs represents a fundamental trade-off between supervision density and scalability. PRMs offer granular, step-by-step guidance that significantly narrows the search space, thereby facilitating faster and more stable policy convergence. 
In contrast, ORMs are implementation-friendly and provide the unconstrained exploration freedom necessary for models to discover non-trivial reasoning paths, such as emergent self-correction, that may deviate from human-annotated trajectories. 
Hence, developing hybrid reward frameworks that synergize the dense oversight of PRMs with the scalable exploration of ORMs remains a critical open challenge in the pursuit of more efficient and cost-effective LRMs.
% In summary, PRMs offer granular guidance and make the models easily to convergence, but the construction of PRMs need considerable annotation.
% ORMs are easy to use, while offer exploration freedom.

% Future directions that effectively synergize their strengths remain a critical open question in LRM training.

\subsection{Test-time Methods} \label{subsec: test-time-methods}

% \jh{Test-time Methods}

Test-time methods aim to increase the compute for LLMs at test time to get more accurate and reliable results without post-training and often lift the LRMs to compare with further post-training \cite{snell_scaling_2024}.
We classify the test-time methods into Parallel, Sequential, and Hybrid methods, following \citet{snell_scaling_2024,force-decode-s1}.

\begin{table*}[t]
\centering
\small
\renewcommand{\arraystretch}{1.25} % 稍微增加行高，让表格呼吸感更好
\setlength{\tabcolsep}{5pt}        % 调整列间距

\begin{tabularx}{\textwidth}{
    l                                  % Category (左对齐，用 multirow 合并)
    >{\raggedright\arraybackslash}p{0.28\textwidth} % Method (固定宽度，防止引用太长挤压其他列)
    c                                  % Extra Model (居中)
    >{\raggedright\arraybackslash}p{0.12\textwidth} % Scoring Signal
    >{\raggedright\arraybackslash}X    % Decision / Control (自动填充剩余空间)
}
\toprule
\textbf{Category} & \textbf{Method} & \textbf{Extra Model} & \textbf{Signal} & \textbf{Decision Strategy} \\
\midrule

% --- Parallel Group ---
\multirow{4}{*}{\textbf{Parallel}} 
& Self-Consistency \cite{selfconsistency} 
& \xmark 
& None 
& Majority vote over sampled paths \\

& Shortest \cite{dinardi2025the} 
& \xmark 
& Length 
& Select the shortest solution \\

& Best-of-\(N\) \cite{brown2024largelanguagemonkeysscaling,ovm-1} 
& \cmark 
& ORM 
& Select top-1 via outcome score \\

& Learned Aggregators \cite{qi2025learning} 
& \cmark 
& Learned 
& Merge candidates into new output \\

\midrule 

% --- Sequential Group ---
\multirow{3}{*}{\textbf{Sequential}} 
& Self-Refine \cite{zeng2025evolvingllmsselfrefinementcapability,shi2025ssrsocraticselfrefinelarge} 
& \xmark 
& Feedback 
& Generate $\to$ Self-feedback $\to$ Revise \\

& Self-Critique \cite{self-refine,xu2025doublecheckerenhancingreasoningslowthinking} 
& \xmark 
& Critique 
& Iterative critique and revision \\

& Least-to-Most \cite{zhou2023leasttomost,yang2024buffer} 
& \xmark 
& Prompt 
& Decompose into ordered sub-steps \\

\midrule 

% --- Hybrid Group ---
\multirow{3}{*}{\textbf{Hybrid}} 
& Tree-of-Thought (ToT) \cite{tot,prm-1} 
& \cmark 
& PRM 
& BFS/DFS with state evaluation \\

& Guided Beam Search \cite{guided-beam-search-2,wang_math-shepherd_2024} 
& \cmark 
& PRM 
& Beam expansion with pruning \\

& MCTS \cite{mcts-1,mcts-2} 
& \cmark 
& Value Func 
& Exploration via backpropagation \\

\bottomrule
\end{tabularx}
\caption{Taxonomy of test-time scaling methods. We categorize methods into Parallel (independent sampling), Sequential (iterative refinement), and Hybrid (structured search) paradigms.}
\label{tab:testtime-methods}
\end{table*}
\FloatBarrier

\paragraph{Parallel Decoding}
Parallel test-time methods operate by generating multiple candidate reasoning trajectories and employing a selector or aggregator to derive the final answer. Consequently, the design of the selection mechanism is central to these paradigms.
The most straightforward approach is majority voting, popularized by Self-Consistency \cite{selfconsistency}.
More sophisticated strategies utilize outcome-based scoring or re-ranking, such as Best-of-
$N$ selection and weighted voting, often guided by Outcome Reward Model (ORM) signals \cite{ovm-1,ovm-2,ovm-3,brown2024largelanguagemonkeysscaling,wang_math-shepherd_2024}.
Beyond heuristic rules, recent research explores learned aggregators that directly synthesize a final prediction from the sampled set \cite{qi2025learning}, or frames parallel scaling as population optimization through iterative evolutionary dynamics \cite{zhang2025populationevolveparallelsamplingevolutionary}.
Notably, even lightweight heuristics, such as selecting the shortest solution, have demonstrated competitive performance \cite{dinardi2025the}.
Ultimately, these methods leverage the diversity of candidate solutions to enhance both reasoning accuracy and robustness.

% Parallel test-time methods generate multiple candidate reasoning trajectories and then apply a selector or aggregator to select the best condidate or produce a refined prediction as the final answer.
% Hence, the design of the condadate selector is the core of the parallel decoding paradigms.
% A simple yet effective aggregator is majority voting, as exemplified by Self-Consistency \cite{selfconsistency}.
% More general strategies rely on outcome-based scoring or re-ranking, including best-of-$N$ selection and weighted voting, often guided by ORM signals \cite{ovm-1,ovm-2,ovm-3,brown2024largelanguagemonkeysscaling,wang_math-shepherd_2024}.
% Beyond rule-based aggregation, recent work investigates learned aggregators that directly combine a set of sampled solutions \cite{qi2025learning}.
% Parallel test-time scaling can also be framed as optimizing a population of candidates through iterative selection dynamics \cite{zhang2025populationevolveparallelsamplingevolutionary}.
% Interestingly, even lightweight heuristics such as selecting the shortest solution can achieve competitive performance \cite{dinardi2025the}.
% Overall, these methods exploit the diversity of multiple solutions to improve accuracy and robustness.

\paragraph{Sequential Decoding}
Sequential test-time methods allocate test-time compute along a single evolving trajectory.
A representative pattern is Self-Refine\cite{10.5555/3666122.3668141}, which alternates generation, self-feedback, and revision. Recent work further strengthens this process through training or step-level verification\cite{zeng2025evolvingllmsselfrefinementcapability,shi2025ssrsocraticselfrefinelarge}.
Reflexion\cite{shinn2023reflexion} extends self-refinement from single outputs to long-horizon agent behavior by storing natural-language reflection summaries of failed trials as episodic memory, which is then reused in subsequent attempts as a lightweight form of verbal reinforcement learning.
To reduce ungrounded internal states, ReflAct further enforces explicit goal-state reflection by maintaining a structured belief state and assessing goal alignment before acting \cite{kim2025reflactworldgroundeddecisionmaking}.
Beyond post-hoc correction, Least-to-Most prompting improves the trajectory by decomposing a hard problem into an ordered sequence of easier subquestions, optionally reusing cached reasoning templates \cite{zhou2023leasttomost,yang2024buffer}.
Nevertheless, evidence from CorrectBench suggests that self-correction gains are task- and difficulty-dependent and can introduce unnecessary latency or even hurt simpler cases \cite{tie2025llmscorrectthemselvesbenchmark}, while the dynamics also differ between open-ended generation and multiple-choice settings \cite{rahmani2025selfcorrectinglargelanguagemodels}.

\paragraph{Hybrid Methods}
Hybrid methods combine sequential exploration with parallel proposal generation, using value or verification signals to guide branching and pruning.
Guided beam search incorporates self-evaluation or outcome-supervised value models to rank partial candidates during decoding, improving planning in mathematical reasoning \cite{guided-beam-search-2,ovm-guided-beam-search-1}.
Tree-structured search explicitly expands and evaluates intermediate ``thought'' nodes to enable lookahead and backtracking \cite{tot}, while related tree-oriented aggregation such as ToM uses hierarchical composition to preserve coherence over long contexts \cite{guo-etal-2025-tom}.
More broadly, search can be paired with diversity-enhancing generation to populate candidate branches \cite{ruan-etal-2025-g2} and with value-guided MCTS-style exploration \cite{mcts-1,mcts-2}.
In practice, these methods often rely on process reward models to score intermediate steps and select the most promising branches \cite{prm-1,wang_math-shepherd_2024,prm-2-qwen}, thereby yielding high-quality reasoning trajectories.

\paragraph{Summary}
Previous work found that the potential of LLMs is not fully reached, and the test-time methods are effective techniques to \textbf{approach the upper bound of LLMs}.
\citet{brown2024largelanguagemonkeysscaling} found that LLaMA-3-8B-Instruct can achieve 98.44\% accuracy with 10,000 times repeated sampling and self-consistency, while only 82.9\% with 100 samples.
The state-of-the-art LRMs, e.g., o1 and R1, all exhibit inherent test-time scaling abilities, such as self-refinement, back-tracing, and thought-switching behaviors \cite{yeo_demystifying_2025} in their extensive intermediate steps.
Moreover, repeated sampling could further improve the performances of R1 \cite{deepseek-r1} and R1-distilled LLMs.
In particular, DeepSeek-R1-Distill-Qwen-14B \cite{ovm-3} achieves 80\% of accuracy on AIME24 \cite{aime} by applying majority voting on 64 samples, with only 69.7\% accuracy of pass@1.
Moreover, \citet{snell_scaling_2024} found that the test-time methods are even more effective compared with additional training in easy and medium problems, while this is not the case on difficult problems.

\section{Challenges Towards Reasoning Economy } \label{sec:challeng}

% \begin{figure*}[!t]
%     \centering
%     \includegraphics[width=0.5\linewidth]{Figs/pt-bad-behavior.pdf}
%     \caption{An illustration of Overly Cautious and Fake Thinking Behaviors of LRMs. Circles: steps; Solid circles: answers.
%     Overly cautious LLMs may repeatedly verify even after finding the correct answer.
%     Fake thinking LLMs may show behaviors like self-refinement or back-tracing. Yet, these behaviors are often superficial, not driven by a genuine awareness of needing to improve the answer.}
%     \label{fig:pt-bad-behavior}
% \end{figure*}

In this section, we primarily examine the challenges in LRMs that impair the reasoning economy.
We first analyze the Inefficient Model Behaviors caused by post-training methods, including Overly Cautious and Superficial Reasoning, which not only impair the LRMs' performance but also lead to significant computation waste.
Then, we reveal the causes of waste of computation during test-time, the Inefficient Budget Allocation,
to exhibit the importance of adaptive test-time setting to reduce computation while achieving higher performance.

\begin{table*}[t]
  \centering
  \small
  % 增加行高，让表格不那么拥挤
  \renewcommand{\arraystretch}{1.3} 
  \setlength{\tabcolsep}{6pt}
  
  % 定义列格式：
  % 第一列和第二列稍窄，第三列宽
  % 这里使用调整后的 X 列 (总和必须等于 X 列的数量，这里是3)
  \begin{tabularx}{\textwidth}{
      >{\raggedright\arraybackslash\hsize=0.6\hsize}X 
      >{\raggedright\arraybackslash\hsize=0.9\hsize}X 
      >{\raggedright\arraybackslash\hsize=1.5\hsize}X 
    }
  \toprule
  \textbf{Category} & \textbf{Specific Behavior} & \textbf{Mechanism Analysis} \\
  \midrule
  
  % 第一组
  \multirow{4}{=}{\textbf{Overly}\\ \textbf{Cautious}} & 
  \textbf{Length Bias} \newline \cite{singhal2024a,chen2024odin,park-etal-2024-disentangling} 
  & \textbf{Cause:} Reward models inherently prefer longer responses. \newline
    \textbf{Symptom:} Outputs are verbose with low information density (low entropy). \\
  \cmidrule(l){2-3} % 使用局部短横线或者 \cdashline{2-3}
   & 
  \textbf{Overthinking} \newline \cite{over-cau-1,longer-is-not-better,lost-in-middle}
  & \textbf{Cause:} Sparse rewards discourage risk-taking. \newline
    \textbf{Symptom:} Excessive self-verification steps on simple tasks. \\
    
  \midrule[0.5pt] % 如果不喜欢实线，可以用 \hdashline
  
  % 第二组
  \multirow{4}{=}{\textbf{Superficial}\\ \textbf{Reasoning}} & 
  \textbf{Fake Thinking} \newline \cite{decp-behavir-1-underthink,self-critique-bc-2}
  & \textbf{Cause:} Underlying biases in SFT data or reward hacking. \newline
    \textbf{Symptom:} Shallow imitation of reasoning structures (e.g., "Let's think") without logical validity. \\
  \cmidrule(l){2-3}
   & 
  \textbf{Underthinking} \newline \cite{decp-behavir-1-underthink,RB}
  & \textbf{Cause:} Inadequate exploration incentives. \newline
    \textbf{Symptom:} Giving up too early or failing to explore complex branches on hard cases. \\
    
  \bottomrule
  \end{tabularx}
  \caption{A compact taxonomy of inefficient reasoning behaviors impacting the reasoning economy. We categorize these behaviors into structural redundancy (Overly Cautious) and qualitative failure (Superficial Reasoning).}
  \label{tab:challenges-ineff-behaviors}
\end{table*}

\FloatBarrier

\subsection{Inefficient Model Behaviors from Post-training} \label{subsec:ineff-pt}

\begin{wrapfigure}{r}{0.4\textwidth}
    \centering
    \includegraphics[width=\linewidth]{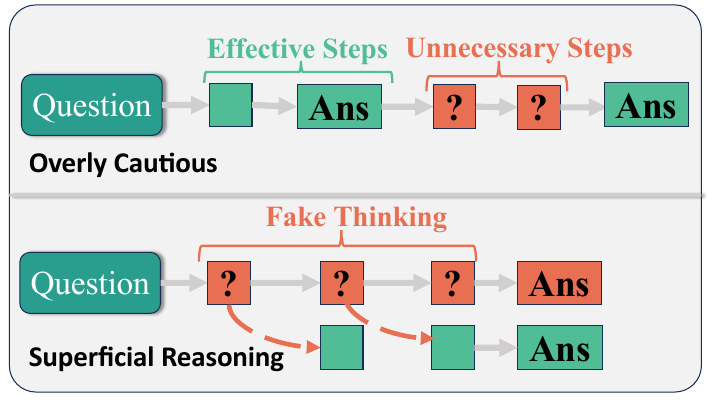}
    \caption{Illustration of \textbf{Overly Cautious} and \textbf{Superficial Reasoning} Behaviors of LRMs. 
    Overly cautious LLMs may repeatedly verify even after finding the correct answer.
    Fake thinking LLMs may show behaviors like self-refinement or back-tracing. Yet, these behaviors are often superficial, not driven by a genuine awareness of needing to improve the answer.}
    \label{fig:pt-bad-behavior}
    \vspace{-3mm} 
\end{wrapfigure}

LRMs are post-trained to align with human preferences and improve reasoning abilities.
RL methods play a key role in the post-training stage for LRMs.
However, RL optimization relies on reward models (RMs) that are inherently imperfect, primarily due to unreliable human preference annotations \cite{singhal2024a,RL-scaling}.
Consequently, over-optimizing based on these flawed RMs can negatively impact the overall capabilities of LLMs.
This situation exacerbates the risk of reward hacking \cite{reward-hack-1}, where models exploit loopholes in the reward function to maximize scores without genuinely aligning with the underlying human preferences.
As a result, LLMs exhibit over-optimization \cite{RL-scaling}, appearing to meet performance expectations during evaluation, but lacking a true understanding.
Next, we will analyze previous research and summarize two notable behaviors that directly impact the reasoning efficiency of LLMs: Overly Cautious and Superficial Reasoning.

One of the most prominent issues is \textbf{Overly Cautious}, where LLMs tend to generate longer responses that contain much redundant content, due to their hacking to maximize their reward scores.
Previous research has shown that LLMs trained with RL tend to produce longer responses \cite{len-grow-1, len-grow-2} compared to those trained through SFT.
As a result, several studies have aimed to answer two key research questions: \textit{RQ1: What are the reasons for longer responses?} and
\textit{RQ2: Does the increased length indicate a bias or an enhancement of model capabilities?}
\citet{singhal2024a} discovered that in existing reward model training datasets, longer responses are often preferred (e.g., RLCD \cite{RLCD}: 63.1\%), which leads to a length preference in the RM (\textit{RQ1}).
Hence, the length-biased RM leads the LLMs to generate redundant content with little performance benefits, e.g., too many paraphrased or connection words.
Furthermore, they found that using length as a proxy for reward models can yield performance comparable to that of PPO with RMs (\textit{RQ2}), which further reveals the underlying bias and questions on the RM accuracy.
Other studies have also indicated that RMs struggle to effectively disentangle length bias from response quality during RL training \cite{singhal2024a, chen2024odin, park-etal-2024-disentangling, lu-etal-2024-eliminating}.

\subsubsection{Overly Cautious Behaviors of LRMs}
% \paragraph{Overly Cautious LRMs}
\citet{deepseek-r1} found that an accuracy- and format-based reward function of the RL process for LRMs is effective without the assistance of PRM or RM, and can largely avoid reward hacking.
Thus, the Deepseek R1 exhibits a surprising improvement in complex reasoning tasks.
Experiments \cite{deepseek-r1, team_kimi_2025} suggest that such improvements from that the LRMs experience an "Aha" moment during the RL process: the reasoning path becomes longer and more complex, leading to behaviors such as self-refinement, recognizing and correcting mistakes, breaking down difficult steps, and iterating on alternative approaches(\textit{R1}, \textit{R2}).
Though progress has been made, recent studies have also identified an "overly cautious" phenomenon \cite{S1-S2-survey, over-cau-1, over-cau-2-agent, longer-is-not-better} in LRMs, e.g., with the naming as overthinking, characterized by excessive verification and redundant reasoning after giving the right answer (\textit{R2}).
This behavior stems from the assumption that longer outputs are more likely to contain the correct answer or appear more comprehensive, even when shorter, more concise responses would suffice.
This overly cautious behavior not only results in inefficient token usage but also hampers LLM performance due to cumulated errors and the "lost in the middle" \cite{lost-in-middle, length-impact-on-reasoning}.

\begin{AIbox}{Overly Cautious LRMs}
LRMs exhibit excessive unnecessary verification and redundant reasoning on easy-to-handle questions or meaningless paraphrases and deviations, leading to inefficient token usage and increased computational costs.
\end{AIbox}

% \subsubsection{Algorithm Bias}

\subsubsection{Superficial Reasoning}

Superficial Reasoning behavior refers to instances where LLMs appear to align with human preferences,
but these behaviors either fail to produce tangible outcomes \cite{fake-align-1} or conceal other underlying objectives \cite{fake-align-2-demo-cheating,decp-behavir-2-benchmark-wdct}.
For example, \citet{fake-align-2-demo-cheating} has demonstrated that LLMs may display differential behaviors across various demographic groups, which could only be found by comparing LLMs' responses to different people.
Deceptive behavior is more challenging to detect than length bias or overly cautious behaviors,
thus extensive and meticulous human observations \cite{fake-align-2-demo-cheating, decp-behavir-1-underthink} are needed to identify it.

\paragraph{Fake Thinking LRMs}
In the context of LRMs, recent studies have uncovered a \textbf{Fake thinking} behavior: They tend to generate plausible and fluent reasoning steps, but these trajectories are often unfaithful or lack logical rigor \cite{shen2025faithcotbenchbenchmarkinginstancelevelfaithfulness}.
Some studies \cite{fake-self-refine-1,decp-behavir-1-underthink,self-critique-bc-2} found that LRMs may appear to engage in self-refinement or deliberate analysis, yet the generated chains can be post-hoc rationalizations or spurious reasoning that is difficult to detect from surface form alone \cite{shen2025faithcotbenchbenchmarkinginstancelevelfaithfulness}.
Empirical evidence suggests that such behaviors often lead to little substantial progress: models may oscillate between alternatives, rewrite reasoning without convergence, or abandon high-quality steps early and replace them with ineffective refinements \cite{decp-behavir-1-underthink}.
These deceptive behaviors not only waste test-time compute but can even hurt performance \cite{self-critique-bc-1,self-critique-bc-2}.
Moreover, \citet{anderson2025phdknowledgerequiredreasoning} found that R1 can be poorly calibrated, giving up quickly on some solvable problems while getting stuck ``thinking forever'' on others.
According to the reasoning boundary theory \cite{RB}, fake thinking can also be treated as the meaningless attempts of LLMs while solving problems beyond their capabilities.

\begin{AIbox}{Superficial Reasoning LRMs}
LRMs appear to work towards problem-solving, but their reasoning trajectories can be unfaithful or ineffective, leading to wasted computation and sometimes worse answers.
\end{AIbox}

\subsection{Inefficient Model Usage in Test-time}\label{subsec:ineff-tt}

Though the test-time methods could push the performance of LRMs
further in a training-free way,
the application of test-time methods is often suboptimal \cite
{snell_scaling_2024,length-analysis-on-llms-1,early-stop-3}.
Previous work found that two factors significantly influence test-time performance:
(1) the selection of inference strategy (e.g., sampling, search, verification) \cite{snell_scaling_2024},
and (2) the test-time computation allocated to each question \cite{length-analysis-on-llms-1,early-stop-3}.

\subsubsection{Unreasonable Algorithm Selection}
To employ a test-time method, there are two dimensions to be
decided if we do not need to consider computation limitation: (1) which inference strategy to use and (2) what hyper-parameters to set (e.g., temperature and top-$p$).
However, naive choices can waste computation and even degrade accuracy.
In particular, a one-size-fits-all test-time inference strategy is rarely effective; the optimal choice depends on the task, model family, and problem difficulty (and often the compute budget) \cite{not-alog-is-all,agarwal2025artscalingtesttimecompute}.

For the choice of inference algorithm, \citet{not-alog-is-all} found that there is not a single inference algorithm that suits all of the tasks.
To be more specific, \citet{vote-lead-to-worse} found that majority voting in LLMs improves accuracy on simple problems but degrades performance on complex ones as votes increase.
Similarly, \citet{snell_scaling_2024} concluded that search-based methods outperform parallel methods on harder problems.
\citet{veri-needed-setlur_scaling_2025} argue that verifiers play a central role in effective test-time scaling.
However, recent evidence suggests that under many practical inference budgets, scaling solution generation (e.g., Self-Consistency) can be more compute-optimal than scaling verification \cite{singhi2025solveverifycomputeoptimalproblem}.
As for the parameter setting of the inference algorithm, recent work  \cite{daptive_temp} found that a lower temperature of sampling is more suitable for reasoning tasks, while creative tasks need a higher temperature.
Moreover, reasoning capabilities can be highly sensitive to subtle implementation choices such as decoding parameters, random seeds, and prompt formatting \cite{hochlehnert2025soberlookprogresslanguage}.
The above studies emphasize \textbf{the need for algorithmic adaptability based on task complexity}.

\subsubsection{Unreasonable Computation Allocation}
As mentioned before, though scaling the computation could bring consistent performance benefits, scaling LLaMA-3-8B-Instruct from generating 100 samples to 10,000 samples is often unacceptable for a simple question.
However, a more complex problem is worth a high computation budget, while a small computation budget could lead to suboptimal accuracy.
Recent studies propose the Reasoning Boundary \cite{RB} of LRMs, in which they found that middle-complexity problems need more computation.
For sequential inference, several studies found that longer solutions with more self-refinement are not necessarily better \cite{over-cau-1,over-cau-4-overthinking,longer-is-not-better,length-analysis-on-llms-1}; instead, there is an optimal refinement length that increases with problem difficulty \cite{over-cau-1}.
\citet{snell_scaling_2024} further found that compute-optimal allocation varies with problem complexity, and \citet{domain-optimal-len} reported that optimal length distributions can differ across domains.
Moreover, reasoning models can be poorly calibrated on easy questions, generating excessive tokens with little accuracy gain, and decoding-time interventions can mitigate such overthinking \cite{pu2025thoughtterminatorbenchmarkingcalibratingmitigating}.
Hence, inadequate computation can lead to wrong answers, while excessive computation may waste tokens and even impair performance.
Above all, previous work empirically found that there is not a one-for-all computation setting for all tasks and samples varied in complexity.
They emphasize \textbf{the importance of adaptive computation allocation based on task complexity}.

% \begin{table*}[ht]
% % \small
% \centering
% \begin{tabular}{llcc}
% \toprule
% \textbf{Method} & \textbf{Type} & \textbf{Regulated Behavior}  & \textbf{Stage}  \\
% \midrule
% LIMO (\citeyear{limr}) & Data & Both  & SFT\\
% S1 (\citeyear{force-decode-s1}) & Data & Both  & SFT \\
% Cosine (\citeyear{yeo_demystifying_2025}) & Algorithm; \textit{Long2short RL} & Both  & RL\\

% Kimi (\citeyear{team_kimi_2025}) & Algorithm; \textit{Long2short RL} & Both  & RL\\
% MRT (\citeyear{qu_MRL_optimizing_2025}) & Algorithm; \textit{Long2short RL} & Both & RL \\
% LCPO (\citeyear{length-control-RL-1}) & Algorithm; \textit{Adaptive Budget-aware Tuning} & Overly Cautious  & RL\\

% Coconut (\citeyear{hao2024traininglargelanguagemodels}) & Algorithm; \textit{CoT Compression} & -  & SFT\\
% o1 (\citeyear{o1}) & Architecture; \textit{Sys-1 \& Sys-2 Cooperation} & - & -  \\
% CITER (\citeyear{zheng2025citer}) & Architecture; \textit{Sys-1 \& Sys-2 Cooperation} & -  & -\\

% \bottomrule
% \end{tabular}

% \caption{Part of works that aims to regulate the behaviors of reasoning LLMs. }
% \end{table*}

\section{Optimization for Reasoning Economy\\ \quad \quad \quad \quad \textit{part-1: Post-training}} \label{sec:solu-pt}

\begin{figure*}
    \centering
    \includegraphics[width=1\linewidth]{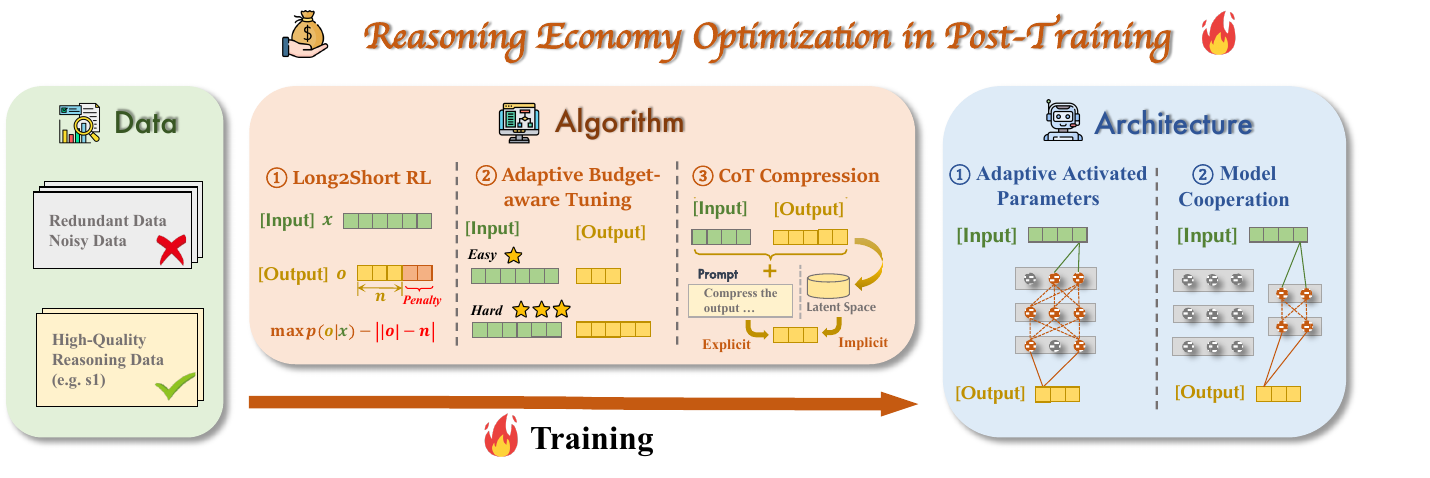}
    \caption{Post-training Methods Optimization for Reasoning Economy.}
    \label{fig:post-train-op}
\end{figure*}

In this section, we provide an overview of post-training optimization methods aimed at enhancing reasoning economy in LLMs.
These methods focus on regulating LLM behaviors to improve efficiency and reduce unnecessary computational overhead from Data, Algorithm, and Architecture perspectives.

% \input{Tabs/table-posttrain-optimization}
% \FloatBarrier

\subsection{Data} \label{subsec:solu-pt-data}

\paragraph{High Quality Data Construction}
By explicitly encoding desired reasoning patterns and behaviors, high-quality post-training data can guide LLMs toward more advanced and effective reasoning \cite{lima,limr,o1-journey,force-decode-s1}.
\citet{o1-journey} utilized small-scale long-thought datasets sampled via test-time scaling \cite{mcts-1,mcts-2} and found that the resulting models exhibit clear long-thought reasoning patterns.
Similarly, \citet{force-decode-s1} demonstrated that only 1,000 high-quality and diverse SFT samples can produce competitive LRMs comparable to advanced models such as o1-preview, highlighting the importance of data quality, diversity, and difficulty.
Recent studies further suggest that data quality is also shaped by selecting a small set of high-impact examples that serve as effective cognitive templates for eliciting reasoning \cite{ye2025limoreasoning,limr,wang2025reinforcementlearningreasoninglarge}.
For example, \citet{ye2025limoreasoning} showed that strong reasoning can emerge from only a few carefully designed demonstrations, while \citet{limr} found that a strategically selected subset of RL data can outperform much larger training sets.
Moreover, \citet{wang2025reinforcementlearningreasoninglarge} reported that even one-shot RLVR can yield substantial gains, highlighting the importance of sample selection and data efficiency in post-training.

\subsection{Algorithm} \label{subsec:solu-pt-alg}

Recent research has focused on three key strategies: optimizing reward structures to balance response length and quality (Long2short RL), enforcing explicit budget control during post-training and inference (Adaptive Budget-aware Tuning), and developing compression techniques to eliminate redundant reasoning steps while preserving accuracy (CoT Compression). These approaches collectively aim to enhance inference efficiency by either directly guiding models to generate concise responses or restructuring reasoning processes for more compact representations.

\subsubsection{Long2short RL}
To address the inefficiencies caused by length bias in RL-tuned LLMs, researchers have proposed various reward design improvements.
For example, \citet{singhal2024a} explored intuitive approaches, such as increasing the KL coefficient, applying length penalties to reward model scores, and discarding overly long rollouts. Despite these efforts, RL-tuned models still tend to produce longer responses than SFT models.
Recent work suggests that such verbosity does not always reflect deeper reasoning, but can also arise as an RL optimization artifact when models frequently produce incorrect answers, especially when training is dominated by unsolvable problems \cite{fatemi2025concisereasoningreinforcementlearning}.

To address this, \citet{team_kimi_2025} proposed several approaches to transform lengthy and unnecessary reasoning processes into concise and accurate ones.
They investigated strategies like model merging, shortest rejection sampling, and DPO optimization for shorter responses. 
They also introduced the \textbf{long2short RL}, which uses a normalized reward model across multiple responses, significantly reducing output length while maintaining reasoning quality.
The following section explores the long2short RL approach designed to improve the efficiency and effectiveness of RL-tuned LLMs.

\paragraph{Quality-Length Reward Disentangle}
One approach to mitigate length bias is to develop more sophisticated reward models that can better distinguish between response quality and length \cite{park-etal-2024-disentangling,chen2024odin}.
These methods employ a multi-head reward model on shared feature representations: one head is explicitly supervised to predict sequence length, while the other is optimized for quality. This architectural separation forces the model to isolate length information, ensuring the quality reward signal remains independent of the output length. 
\citet{li2025drpoefficientreasoningdecoupled} propose the DRPO, decoupling the length-based optimization objectives for correct and incorrect rollouts, thereby preventing negative samples from interfering with the learning of positive ones.
Investigating the correlation between accuracy and CoT length, \citet{liang2025deepcompressdualrewardstrategy} proposes an adaptive strategy that aligns reasoning depth with task complexity, encouraging concise responses for simple queries while reserving longer reasoning chains for difficult problems, thereby optimizing the efficiency-accuracy trade-off.

\paragraph{Length Penalty or Normalization}
\citet{simpo} proposed a simple length normalization way for DPO \cite{DPO} to eliminate the influence of length, which has been proven quite effective in alleviating the length bias \cite{over-cau-4-overthinking}.
\citet{yeo_demystifying_2025} utilizes a cosine reward to incentivize different length scaling behaviors to regularize the model behavior, eliminating length bias.
\citet{over-cau-4-overthinking} observed that LRMs often provide multiple correct answers within a single solution.
Leveraging this pattern, they split these solutions into multiple shorter ones and constructed preference pairs, with the first short and correct answer as the preferred solution.
They optimized the LLMs using DPO and SimPO, finding SimPO \cite{simpo} to be more effective, reducing response lengths by 30\% to 40\%.
Beyond uniform length penalties, recent work proposes difficulty-aware penalties that encourage conciseness on easy questions while preserving sufficient reasoning depth on harder ones \cite{ling2025fasteasydeephard}.

\paragraph{Procedure Reward}
Rule-based reward \cite{deepseek-r1}, i.e., accuracy-based reward, has been proved to be quite effective, and less likely to suffer from reward hacking \cite{reward-hack-1}.
However, a recent study \cite{qu_MRL_optimizing_2025} highlights a limitation of accuracy-based rewards in shaping LLM reasoning: they fail to ensure efficient utilization of scaled computation during inference.
To address this, the authors propose the Meta Reinforcement Finetuning (MRT) by using \textit{cumulative regret} as a reinforcement learning objective, encouraging LLMs to make incremental progress with each token, to provide clear procedure instructions.
This approach was shown to enable stable performance improvements as computational resources scale.

\subsubsection{Adaptive Budget-aware Tuning}

Budget-aware tuning aims to align LRMs with explicit token budgets (e.g., reasoning trace length), enabling controllable trade-offs between accuracy and inference cost.
Compared with purely test-time budget prediction and constrained generation (\S~\ref{subsec:solu-tt-input}), this line of work primarily improves \emph{intrinsic} budget following via post-training.

\paragraph{Budget Specification}
A common starting point is to expose models to explicit budget signals at inference time, such as formulating the procedure as an instruction-following problem, by specifying a target length in the prompt \cite{yuan2024-length-cons-sft}.
More recent work conditions the model on more structured budget interfaces, e.g., inserting control tokens that indicate remaining budget during generation \cite{wen2025budgetthinkerempoweringbudgetawarellm}, or providing discrete user-controllable reasoning modes associated with different budget tiers \cite{zhao2025saberswitchablebalancedtraining}.

\paragraph{Budget-aligned Post-training}
% Beyond prompting, models can be post-trained to follow budget constraints more reliably while preserving task performance.
% \citet{yuan2024-length-cons-sft} constructs SFT data with explicit length specifications to improve length controllability, and \citet{length-control-RL-1} further adopts RL to jointly optimize accuracy and length control across different budgets.
% To better maintain capability under multi-tier budgets, \citet{lyu2025hierarchicalbudgetpolicyoptimization} proposes hierarchical budget policy optimization, which structures exploration and rewards across budget levels to mitigate collapse toward overly short solutions.
% \citet{BudgetThinker} proposes to utilize the special token which indicate the LRMs the remaining budget, and post-training the models to present budget-aware behaviors accordingly.
% \citet{li2025steeringllmthinkingbudget} utilize the constructed Gamma distribution to guide the next-token prediction, to achieve reliable budget-aware LRMs.
% Above all, 

To instill a more intrinsic awareness of resource limits, recent research has moved toward post-training models for budget compliance. 
While early efforts focused on supervised instruction-following with length constraints \cite{yuan2024-length-cons-sft}, subsequent works integrated RL to balance the trade-off between reasoning depth and compute budgets \cite{length-control-RL-1}. 
To address the challenge of capability collapse under tight budgets, \citet{lyu2025hierarchicalbudgetpolicyoptimization} proposed a hierarchical reward structure that preserves reasoning quality across different expenditure levels. Furthermore, researchers have introduced budget-aware architectural cues, such as remaining-budget tokens \cite{wen2025budgetthinkerempoweringbudgetawarellm} or probability steering via Gamma distributions \cite{li2025steeringllmthinkingbudget}, to achieve finer-grained control. 
Above all, these advancements demonstrate that budget awareness can be effectively encoded into the model's parameters, transforming LRMs into resource-conscious ones that optimize the utility-to-cost ratio at the foundational level.

\paragraph{Adaptive Allocation and Gating}
Since not all instances require the same reasoning depth, a key direction is to make budget usage adaptive.
\citet{yan2025drqadynamicreasoningquota} trains models to allocate a reasoning quota conditioned on perceived difficulty, reducing unnecessary computation on easy queries while retaining depth for hard ones.
Relatedly, hybrid-reasoning models learn to decide when to ``think'' and when to answer directly \cite{jiang2025think}.
These approaches also help mitigate the token elasticity phenomenon, where overly strict constraints can trigger counterproductive behaviors and higher token costs \cite{0shot-sft-length-1}, by favoring flexible, instance-dependent budget allocation over rigid length targets.

\subsubsection{CoT Compression}

It is evident that longer CoT suffers from token redundancy and inference latency, which is not desired in practice. Thus, several studies try to identify important tokens and eliminate unnecessary tokens or reasoning steps, therefore enhancing the inference economy. These methods can be broadly categorized into two categories: 1) \textit{explicit compression} that directly enforces the model to generate more concise reasoning by fine-tuning on carefully curated datasets or providing specific demonstrations.; and 2) \textit{implicit compression} that maps multiple reasoning tokens or steps into a continuous space to achieve a more compact representation.

\paragraph{Explicit Compression} The first step of compression is to identify key tokens \cite{lee_how_2025} or step in the reasoning processing, while most methods rely on perplexity as a primary metric, using it as a proxy to determine the importance of individual tokens or steps.
Some of the methods refine the supervised fine-tuning dataset, replacing lengthy reasoning with identified key reasoning steps as target outputs \citep{xia2025tokenskip}.
Another approach uses key reasoning steps as demonstrations for in-context learning, guiding LLMs to generate only essential steps during inference while minimizing unimportant tokens~\citep{cui2025stepwise}.
A key limitation of these methods is the potential loss of coherence and a comprehensive understanding of the full reasoning process.
While the full CoT can be reconstructed using additional LLMs, the recovered version may not fully align with the original, leading to inconsistencies or missing nuances and additional inference costs.

\paragraph{Implicit Compression} Besides explicit compression, another line of work argues that the explicit language space (i.e., tokens) may not always be optimal for reasoning, as most word tokens are primarily for textual coherence and not essential for reasoning~\citep{deng2023implicitchainthoughtreasoning, hao2024traininglargelanguagemodels}.
The earlier work focuses on the \textit{implicit} chain-of-thought approach, which compiles explicit CoT reasoning of the teacher model into a student model that directly produces the final answer to develop more efficient reasoning~\citep{deng2023implicitchainthoughtreasoning, deng2024explicitcotimplicitcot}.
Another representative work is Coconut (Chain of Continuous Thought)~\citep
{hao2024traininglargelanguagemodels}, which feed hidden state instead of
specific token back to the LLM as the subsequent input embedding, allowing
the model to encode multiple alternative next reasoning steps in the
continuous space and thus reducing the token cost, followed by~\citet
{cheng2024compressedchainthoughtefficient}.
More recently, CoLaR introduces a latent head to dynamically compress reasoning chains in latent space and exposes a controllable compression factor at inference time \citep{tan2025thinksilentlythinkfast}.
Recent advances in memory-efficient architectures, such as Anchor-based LLMs (AnLLMs)~\citep{pang-etal-2024-anchor}, further highlight the inefficiency of token-based representations by demonstrating that compressing sequence information into anchor tokens can significantly reduce memory usage while maintaining accuracy, followed by~\citet{zhang2025lightthinker}.
To reduce the need to fully finetune the models, recent studies use an additional projection module to inject the compressed continuous states of the source model into the target model~\citep{xu2025softcotsoftchainofthoughtefficient}.

\subsection{Architecture}\label{subsec:solu-pt-arc}

One approach to enhancing efficiency is reducing active model parameters, requiring adjustments in model or system architecture.
Two key strategies are integrating System-1 and System-2 thinking and adaptively utilizing model depth. System-1 and System-2 cooperation enables dynamic selection between fast, intuitive reasoning and slower, deliberate processing, optimizing efficiency.
Meanwhile, adaptive parameter activation optimizes model depth and resource allocation during inference, balancing performance and computational cost.

\subsubsection{System-1 and System-2 Cooperation}

Drawing inspiration from human cognition, several studies propose different methods to dynamically switch from slow, conscious, deliberate reasoning (System 2) to fast, automatic, intuitive thinking (System 1), therefore achieving a better balance between fast and accurate thought. Specifically, there are different paths towards this goal.

\paragraph{Single-Model Routing}

It is straightforward to empower one model with abilities to switch between fast (system-1) and slow (system-2) inference methods according to difficulty signals, thereby optimizing both efficiency and effectiveness.
On the one hand, the choice can be left to the user to determine the extent of cognitive effort exerted by the model.
For instance, OpenAI has equipped the O1 \cite{o1} with three distinct thinking modes: low, middle, and high.
The high thinking mode delivers accurate yet lengthy responses tailored for extremely challenging problems, while the low mode provides swift responses that are satisfactory.
However, this direction has not been sufficiently explored thus far.
How to train models to possess diverse modes of thinking and how to evaluate the capability gap between different modes \cite{early-giveup} of thinking in models remain underexplored areas.

On the other hand, the model can be designed to autonomously determine the appropriate level of cognitive effort based on the complexity of the task at hand \cite{wang2025mixllmdynamicroutingmixed,ong2025routellmlearningroutellms,pan-etal-2024-dynathink,saha2024system,ding_hybrid_2024}.
This autonomous selection process can be facilitated by incorporating a mechanism that assesses the difficulty of the input and dynamically adjusts the inference method accordingly.
\citet{ding_hybrid_2024} leverages a binary router to assign queries to smaller or larger based on the difficulty and expected quality.
\citet{pan-etal-2024-dynathink} use consistency verification and reasoning steps as complexity signals to determine which questions require slow, deliberate thinking and which do not. Similarly, \citet{saha2024system} curates a dataset that trains models to decompose problems into sub-goals of varying difficulty levels and adaptively apply different strategies to solve them.

\paragraph{Multi-Model Collaboration} One common approach to accelerating inference is speculative decoding~\citep{xia-etal-2023-speculative,kim2023speculative,leviathan2023fast}, which follows a draft-then-verify paradigm. This method first generates multiple token candidates efficiently and then verifies them in parallel. \citet{xia-etal-2023-speculative} formally analyze this method through a formal study and extensive discussion of both drafting and verification phases, while \citet{kim2023speculative} proposes the BiLD framework, where a smaller model generates text at low cost, and a larger model refines errors in parallel. Further research explores improved drafting and verification strategies using specific reward model~\citep{bachmann2025judge, liao2025rewardguidedspeculativedecodingefficient}, with a broader overview in \citet{xia-etal-2024-unlocking}. In addition, some methods take an agentic approach, dynamically determining when to engage larger models for refinement at different granularity~\citep{lin2023swiftsage, chen2024magicore, zheng2025citer}. Specifically, SwiftSage~\citep{lin2023swiftsage} invokes system 2 thinking for subgoals during inference, while CITER~\citep{zheng2025citer} focuses on token-level refinement when needed.

\paragraph{Knowledge Distillation}
Another approach to achieving a balance between slow and fast reasoning is knowledge distillation, where knowledge from a larger, more complex model (System 2) is transferred to a smaller, more efficient model (System 1).
The distilled smaller model will be capable of advanced reasoning abilities while costing less in computations.
In detail, \citet{yu2024distilling} leverages several self-supervised methods to distill higher-quality outputs from System 2 techniques back into LLM generations without intermediate reasoning token sequences.
More recently, it has been proven that direct distillation from DeepSeek-R1 outperforms applying RL on the base model such as Qwen2.5-32B, setting a new record on the reasoning benchmarks~\citep{deepseek-r1}.
There are also emerging studies exploring distillation from Transformer to lower computation complexity architectures~\citep{paliotta2025thinking}, such as Mamba, achieving much faster inference for large batches and long sequences with the Mamba-based model.

\subsubsection{Adaptive Activated Parameters}
Performance improvements of deep learning models largely benefit from advancements in deep network design \citep{He_2016_CVPR}.
Present LLMs also pay great attention to deep architecture designs.
For models with comparable scales of parameters, increasing model depths often yields more significant benefits than broadening widths \citep{chen2024what,feng2023towards}, especially in mathematical reasoning tasks \citep{ye2025physics}.
However, deepening layers will result in larger model sizes, which means that relatively small models with few layers are hard to resolve complex reasoning tasks.
Additionally, increasing model layers leads to a linear increase in inference computation costs.
From the perspective of model depth, we can enhance the inference capabilities of relatively small-scale LLMs by recurrently utilizing intermediate layers or accelerate the inference of deep LLMs by skipping some intermediate layers.

\begin{figure*}
    \centering
    \includegraphics[width=1\linewidth]{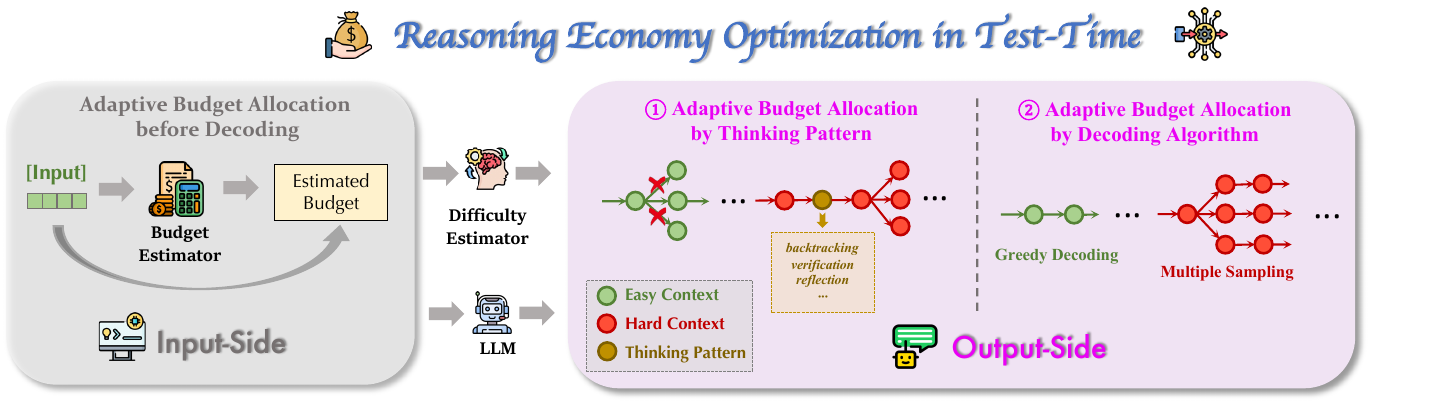}
    \caption{Test-time Methods Optimization for Reasoning Economy. The methods are divided into adding an optimal computation constraint on the input-side, and selecting the best-performing decoding algorithm and controlling computation usage during decoding on the output-side.  }
    \label{fig:test-time-op}
\end{figure*}

\paragraph{Recurrent Layers}
Recurrent mechanisms enable LLMs to perform arbitrarily many computations before emitting a token, which is a simple solution for test-time compute scaling for relatively small LLMs with few layers \citep{geiping2025scalingtesttimecomputelatent}.
Recent work has described depth-recurrent, looped Transformers and studied their potential benefits with careful theoretical and small-scale analysis \citep{pmlr-v202-giannou23a,gatmiry2024loopedtransformerslearnimplement,yang2024looped}.
Specifically, \citet{fan2025looped} established the superior length generalization capabilities of looped Transformers.
\citet{yu2025enhancingautoregressivechainofthoughtloopaligned} proposes a framework enhancing CoT reasoning by combining looped and auto-regressive Transformers.
Recurrently utilizing layers in Transformer-based LLMs enhances reasoning ability on small-scale LLMs.

\paragraph{Dynamic Depth}
Additionally, inspired by efficient reasoning using model pruning and sparse models, recent works validate that not all layers of LLMs are necessary during inference \citep{fan2024layersllmsnecessaryinference}.
In this way, two methods involving dynamic depths of LLMs are included: Early Exit \citep{varshney-etal-2024-investigating,NEURIPS2022_6fac9e31} and Skip Layer \citep{yang-etal-2024-laco,men2024shortgptlayerslargelanguage,kim2024shortenedllamadepthpruning}.
Since simpler tasks require fewer layers, while more complex tasks require more layers of inference, achieving efficient LLM inference then becomes when to stop the inference process adaptively based on the input instance \citep{fan2024layersllmsnecessaryinference}.

\section{Optimization for Reasoning Economy\\ \quad \quad \quad \quad \textit{part-2: Test-time Methods}} \label{sec:solu-tt}

Test-time methods can enhance the performance of LRMs without the need for tuning.
However, as previously noted (\S~\ref{subsec:ineff-tt}), there is significant potential for optimizing test-time usage to achieve greater reasoning efficiency.
In this section, we survey existing methods aiming for reasoning economy by optimizing test-time methods to more economically allocate inference computation as shown in Figure~\ref{fig:test-time-op}.

\subsection{Input-side Optimization} \label{subsec:solu-tt-input}
In this section, we introduce those works that attempt to optimize the usage of LLMs by controlling model behaviors through explicit constraints in the input.

\subsubsection{Adaptive Budget Allocation before Decoding}

The main idea of Adaptive Budget Allocation before Decoding is to decide a computation (thinking-token) budget for each query before generation and then steer the model to respect the budget during generation, enabling instance-level efficiency without changing the decoding algorithm \cite{0shot-sft-length-1, difficulty-aware-budget-pred}.

\paragraph{Budget Prediction}
The basic idea of predicting the budget is to consider the difficulty of the problems to the reasoning LLM.
According to previous works \cite{ce-1,ce-2,liu-etal-2024-llms-learn-uncertainty}, the confidence of the model to solve the question can be estimated and predicted.
As for the computation prediction methods for each question, \citet{0shot-sft-length-1} collected a batch of questions with their optimal budgets and trained a regression model to predict the computation needed for novel prompts.
Meanwhile, \citet{difficulty-aware-budget-pred} considered the difficulty of the questions and allocated budgets similar to those of questions with comparable difficulty.
Beyond external estimators, SelfBudgeter trains the model to explicitly self-estimate a suitable budget before producing the solution, and then optimizes for both budget adherence and efficiency via RL \cite{li2026selfbudgeteradaptivetokenallocation}.
AdaCtrl exposes difficulty-aware budgeting through explicit tags, enabling both adaptive allocation and user override of the intended reasoning depth \cite{huang2025adactrladaptivecontrollablereasoning}.

\paragraph{Budget Constrained Generation}
Given a target budget, a straightforward approach is to specify length limits directly in the prompt \cite{length-cont-in-prompt-1,yuan2024-length-cons-sft}, which LLMs can follow to some extent \cite{yuan2024-length-cons-sft}.
L1 further improves controllability by using RL to optimize accuracy while adhering to user-specified length constraints \cite{aggarwal2025l}.
Instead of hard constraints, Budget Guidance proposes a soft, token-level budget-conditioned generation method that guides generation with a lightweight predictor of remaining thinking length, improving accuracy under tight budgets without fine-tuning the base model \cite{li2025steeringllmthinkingbudget}.

\subsection{Output-side Optimization}\label{sec:solu-tt-output}
In this section, we introduce the work of optimizing the usage of LLMs at the output end of the model.
This includes Constrained Decoding, Adaptive Algorithm Selection, and Adaptive Computation Allocation during Decoding.

\subsubsection{Adaptive Algorithm Selection}
The exploration of adaptive choice of test-time algorithms is quite limited, especially in efficient thinking.
Here, we present several works that adaptively adjust the parameters of test-time algorithms, which could be potentially used to achieve algorithm determination.

\citet{snell_scaling_2024} demonstrates the potential of adaptive method selection, showing that optimal settings can achieve up to 4x greater efficiency compared to the PRM best-of-N approach.
This underscores the benefits of dynamically tailoring test-time strategies to specific tasks.
\citet{daptive_temp} introduces an Adaptive Decoding Layer combined with Latent Preference Optimization (LPO), enabling fine-grained temperature prediction at the token or sample level for more precise generation control.
\citet{chen_flaming-hot_2024} propose using high initial temperatures to enhance response diversity, which expands the search space and increases the likelihood of discovering correct answers.
\citet{wang2025adareasoneradaptivereasoningenables} further proposes an RL-based plugin that adaptively selects a factorized configuration (e.g., temperature, reasoning format, and reasoning steps) for each query, enabling more flexible test-time reasoning.
These approaches collectively emphasize the importance of adaptive mechanisms in optimizing the choice of test-time algorithms, paving the way for more efficient and effective reasoning in LLMs.

\subsubsection{Adaptive Budget Allocation During Decoding}

% \textbf{$\triangle$ Early Stopping}
\paragraph{Early Stopping}
The first line of adaptive budget allocation is early stopping, which aims to terminate generation once additional reasoning is unlikely to improve the final answer \cite{early-stop-1,early-stop-2,early-stop-3,early-stop-4,early-stop-5}.
For example, early stopping can be driven by a model's self-evaluation signal during iterative self-refinement \cite{early-stop-1}.
In parallel sampling settings such as self-consistency, a fixed sample size can be wasteful; thus, several methods stop sampling once consistency reaches a threshold \cite{early-stop-2} or progressively narrow the sampling window and terminate when answers converge within a window \cite{early-stop-3}.
The same convergence principle has also been applied to single-trajectory CoT generation by monitoring provisional answers and exiting once they stabilize \cite{mao2025earlystoppingchainofthoughtslarge, liu2025answerconvergencesignalearly}.

Rather than using a uniform stopping rule, recent work emphasizes instance-dependent stopping criteria that account for both path quality and convergence \cite{early-stop-4, early-stop-5}.
One representative approach combines confidence-based selection of preferred reasoning paths with consistency judgment to decide when to stop \cite{early-stop-5}.
Beyond these heuristics, certainty-guided methods periodically probe the model during reasoning and stop once a target confidence threshold is reached \cite{nogueira2025certaintyguidedreasoninglargelanguage}, while internal-confidence based early-exit methods terminate at reasoning transition points when confidence becomes sufficiently high \cite{yang2025dynamicearlyexitreasoning}.
More principled signals can be derived from reasoning dynamics, e.g., detecting instance-specific ``reasoning completion points'' by monitoring termination-token rank dynamics \cite{wei2026evolutionthoughttrackingllm}.
Finally, REFRAIN formulates when-to-stop as an adaptive control problem, combining a redundancy discriminator with an online controller to adjust stopping thresholds under varying difficulty \cite{sun2025stopenoughadaptiveearlystopping}.

\paragraph{Search with Pruning}
The second line is pruning while searching \cite{pruning-while-dec-1,pruning-while-dec-2,pruning-while-dec-3, pruning-while-dec-4}, no matter the parallel or sequential methods.
The goal is to prune low-quality search branches early while retaining high-quality ones,
thereby saving computational resources.
For sequential test-time methods, RMs are often used to guide the search procedure \cite{guided-beam-search-2, ovm-guided-beam-search-1,wang_math-shepherd_2024} by pruning low-quality branches and rolling out high-quality ones. Guided search could eliminate unnecessary computation usage while improving the accuracy of LRMs.

As for parallel test-time methods, \citet{pruning-while-dec-1} utilizes the self-evaluation abilities of LLMs to prune the low-quality samples during best-of-N sampling.
To expand the search space of best-of-N, \citet{pruning-while-dec-3} set a very large sample window at the beginning of the sampling, large enough to consume all of the GPU memory, then gradually pruned those low-quality and completed the saved samples.
\citet{pruning-while-dec-2} select the top high-quality steps and use these steps to guide and constrain the LLMs to complete these prefixes.

\paragraph{Constrained Decoding}

Identifying suboptimal thinking patterns is often challenging, which makes mitigating them particularly difficult.
Hence, previous works have to utilize human analysis to diagnose the deceptive behaviors of LLMs \cite{decp-behavir-1-underthink,decp-behavir-2-benchmark-wdct,fake-self-refine-1}
and more importantly, to utilize human observation to inspire specific solutions for them \cite{find-decp-and-solve-1}.

For example, for fake self-refine of LLMs, \citet{fake-self-refine-2} and \citet{self-critique-bc-1} suggest using external feedback to assist the self-refine procedure of LLMs.
Other works utilize the human-observed behaviors to design force decoding paradigm, aiming to solve the fake thinking and overly cautious behaviors of LRMs, e.g., enforcing LLMs to adhere to the previous thought \cite{decp-behavir-1-underthink} or performing self-refine at proper time \cite{force-decode-s1}.
\citet{decp-behavir-1-underthink} found that the proposed Thought Switching Penalty approach can encourage LLMs to perform deeper reasoning and reduce unreasonable thought-switching.

\section{Discussion}
\label{sec:dis}

\paragraph{Efficient Multi-modal Reasoning}
Multi-modal large language models (MLLMs) have demonstrated promising capabilities in various multi-modal reasoning tasks~\citep{wang2024exploring,zhangmultimodal}, including but not limited to mathematical reasoning~\citep{wang2024measuring,lu2024mathvista,zhou2025is}, visual question answering (VQA)~\citep{goyal2017making,hudson2019gqa}, and multi-modal dialogue systems~\citep{huang2023sparkles}. Recent advances, exemplified by o1-like and R1-like technologies, have further catalyzed progress in this field~\citep{huang2025vision,meng2025mm,peng2025lmm,zhang2025r1,feng2025videor1,yao2024mulberry}. Current approaches on efficient multi-modal reasoning primarily focus on improvements within MLLMs themselves~\citep{jin2024efficient}, mainly including: i) model architecture optimization (e.g., lightweight vision encoders~\citep{chen2024vitamin}, vision token compression~\citep{guo2024llava}, vision-language projector~\citep{li2023blip}, small language models~\citep{chu2023mobilevlm,zhu2024llava}), and efficient structures~\citep{lin2024moe,qiao2024vl,lin2024boosting}; ii) efficient vision techniques adoption (e.g. Pruning~\citep{yu2022width}, knowledge distillation~\citep{touvron2021training}, and ViT quantization~\citep{li2022q}), among other methods.

However, the evaluation and targeted optimization of efficiency in multi-modal (long-) reasoning remain relatively preliminary. \citet{jiang2025mme} proposes two metrics for measuring the efficiency of multi-model CoT: relevance rate, which calculates the proportion of content contributing to answering, and reflection quality, which evaluates whether reflection steps can effectively correct previous mistakes or present new insights. \citet{xiang2025can} proposes a self-structured COT paradigm that decomposes reasoning into minimal semantic atomic steps to achieve efficient multi-modal reasoning by generating dynamic structures and lengths based on the problem types. Moreover, most of the aforementioned efficient reasoning methods for LLMs can also be effectively applied to MLLMs.

\paragraph{Efficient Agentic Reasoning}
% agent accelerate
The advent of LRMs has also brought substantial performance improvements to AI agents. Using RL training similar to o1, the OpenAI's Deep Research~\citep{deepresearch} utilizes advanced reasoning capabilities to synthesize large amounts of online information, attaining notable performance on the challenging Humanity's Last Exam~\citep{phan2025humanity}. Recent research has increasingly focused on leveraging long reasoning capabilities in agent systems, where cutting-edge implementations integrate these with retrieval, tool augmentation, domain knowledge incorporation, and other auxiliary components to push performance boundaries~\citep{jin2025search,alzubi2025open,gao2025txagent}.

% agent economy
To analyze the limited effectiveness of LRMs in interactive environments, \citet{over-cau-2-agent} presents the first comprehensive empirical study in agentic tasks, analyzes the overthinking in LRMs, including its manifestations, quantification, and impact on different models, and also proposes potential mitigation approaches accordingly. \citet{zhou2025large} proposes the LaRMA framework to explore the necessity of reasoning capabilities in LRMs for agent scenarios, analyzes the performance differences between LRMs and LLMs across various tasks, and provides insights into optimizing agent performance through hybrid LLM-LRM configurations. Some studies also investigate efficient agent training methods~\citep{chen2025atlas} and efficient interaction schemes for multi-agent systems~\citep{zhou2025reso,wang2025agentdropout,zhang2025cut}.
% Additional research efforts explored strategic pruning techniques in multi-agent systems to enhance interaction efficiency~\citep{}.

\paragraph{Evaluation Metrics and Benchmarks}
% \cite{btm-1,btm-2,btm-3}
With the proliferation of long reasoning and the increasing severity of the over-reasoning problem, researchers have recently begun developing specialized benchmarks and metrics to quantitatively measure reasoning efficiency.
To stress-test overthinking and calibration, DNA Bench shows that current LRMs can generate up to 70$\times$ more unnecessary tokens and fail on tasks that non-reasoning models handle efficiently \cite{hashemi_dna_2025}, while ThoughtTerminator introduces DUMB500 to evaluate calibration on extremely easy problems and proposes a training-free decoding method to improve it \cite{pu2025thoughtterminator}.
Beyond benchmarks, specialized metrics have also been proposed: \citet{over-cau-4-overthinking} introduces outcome-based and process-based efficiency metrics and finds that LRMs often overthink on simple problems, with later reasoning contributing little to accuracy and diversity.
Complementary to over-reasoning, underthinking has been quantified via an underthinking metric, revealing frequent premature thought switching in o1-like models that leads to insufficient reasoning depth \cite{decp-behavir-1-underthink}.
Finally, \citet{anderson2025phdknowledgerequiredreasoning} builds a dataset from NPR Sunday Puzzle Challenges to evaluate reasoning with general (non-PhD-level) knowledge, which also surfaces overthinking-like behaviors.

\paragraph{Explainability of LRMs}

The study of the explainability of black-box LLMs has always been a topic of interest \cite{exp-1_2024,exp-2_2024}.
Particularly, LRMs have explored on their own through RL and have demonstrated reasoning abilities on par with those of human PhD students \cite{deepseek-r1,o1,team_kimi_2025,noauthor_qwq-32b_nodate}.
However, the mechanisms underlying their achievement of such performance remain enigmatic.
Current research on LRMs often focuses on their behavior analysis \cite{over-cau-1,over-cau-2-agent, yeo_demystifying_2025}, such as the observation of overly cautious or fake thinking behaviors, and then retraces the post-training algorithms or test-time methods.

Nevertheless, it is essential to focus on how these models work on the inside, probing the internal mechanisms of LRMs \cite{ant-exp-1,ant-exp-2}.
Moreover, more user-friendly and powerful toolkits \cite{exp-too-1-microscope_2025, ant-exp-2} are also critical for research purposes and large-scale human analysis.
This would help us understand the thought patterns of LRMs, identify their faults, and provide directions for further improvement.

\section{Conclusion}

In this survey, we systematically examined the challenges and solutions involved in achieving an inference economy for large reasoning models, emphasizing the urgent need for efficient reasoning mechanisms that balance computational cost with performance. To the best of our knowledge, this is the first comprehensive review that analyzes the underlying causes, observed phenomena, key challenges, and emerging solutions for enabling efficient reasoning in LLMs, offering a structured roadmap and actionable strategies for practical deployment. By anchoring progress in the principles of inference economy, this survey serves not only as a synthesis of current knowledge but also as a call to action for further research in this path, highlighting the importance of developing more sustainable and scalable models that reason not only effectively but also efficiently.

\bibliographystyle{acm/ACM-Reference-Format}
\bibliography{custom}

\end{document}